\documentclass[final,3p,times,twocolumn,sort&compress]{elsarticle}

\usepackage{amssymb}

\usepackage{textcomp}
\usepackage{pgfplots}
\usepackage[disable]{todonotes}
\setuptodonotes{inline}
\def\BibTeX{{\rm B\kern-.05em{\sc i\kern-.025em b}\kern-.08em
    T\kern-.1667em\lower.7ex\hbox{E}\kern-.125emX}}
\usepackage{subfig}
\usepackage{float}
\usepackage{algorithm}
\usepackage{algpseudocode}
\usepackage{amsmath}
\usepackage{tabularx}
\usepackage{tabulary}
\usepackage{stfloats}
\usepackage{booktabs}
\usepackage{multirow}
\usepackage[]{xcolor}
\usepackage{wrapfig} %

\journal{Robotics and Autonomous Systems}

\begin{document}

\begin{frontmatter}

\title{RainSD: Rain Style Diversification Module for Image Synthesis Enhancement using Feature-Level Style Distribution\tnoteref{label1}}
\author[inst1]{Hyeonjae Jeon}

\author[inst1]{Junghyun Seo}
\author[inst1]{Taesoo Kim}
\author[inst2]{Sungho Son}
\author[inst2]{Jungki Lee}
\author[inst1]{Gyeungho Choi}
\author[inst1]{Yongseob Lim\corref{cor1}}
\ead{yslim73@dgist.ac.kr}
\cortext[cor1]{Corresponding author}

\affiliation[inst1]{organization={Daegu Gyeongbuk Institute of Science and Technology},%
            city={Daegu},
            postcode={42988}, 
            country={South Korea}}
\affiliation[inst2]{organization={Korea Automotive Testing and Research Institute},%
city={Hwaseong-si, Gyeonggi-do},
postcode={18247}, 
country={South Korea}}

\begin{abstract}
Autonomous driving technology nowadays targets to level 4 or beyond, but the researchers are faced with some limitations for developing reliable driving algorithms in diverse challenges. To promote the autonomous vehicles to spread widely, it is important to address safety issues on this technology. Among various safety concerns, the sensor blockage problem by severe weather conditions can be one of the most frequent threats for multi-task learning based perception algorithms during autonomous driving. To handle this problem, the importance of the generation of proper datasets is becoming more significant. In this paper, a synthetic road dataset with sensor blockage generated from real road dataset BDD100K is suggested in the format of BDD100K annotation. Rain streaks for each frame were made by an experimentally established equation and translated utilizing the image-to-image translation network based on style transfer. Using this dataset, the degradation of the diverse multi-task networks for autonomous driving, such as lane detection, driving area segmentation, and traffic object detection, has been thoroughly evaluated and analyzed. The tendency of the performance degradation of deep neural network-based perception systems for autonomous vehicle has been analyzed in depth. Finally, we discuss the limitation and the future directions of the deep neural network-based perception algorithms and autonomous driving dataset generation based on image-to-image translation based.

\end{abstract}

\begin{graphicalabstract}
	\includegraphics[width=1\hsize]{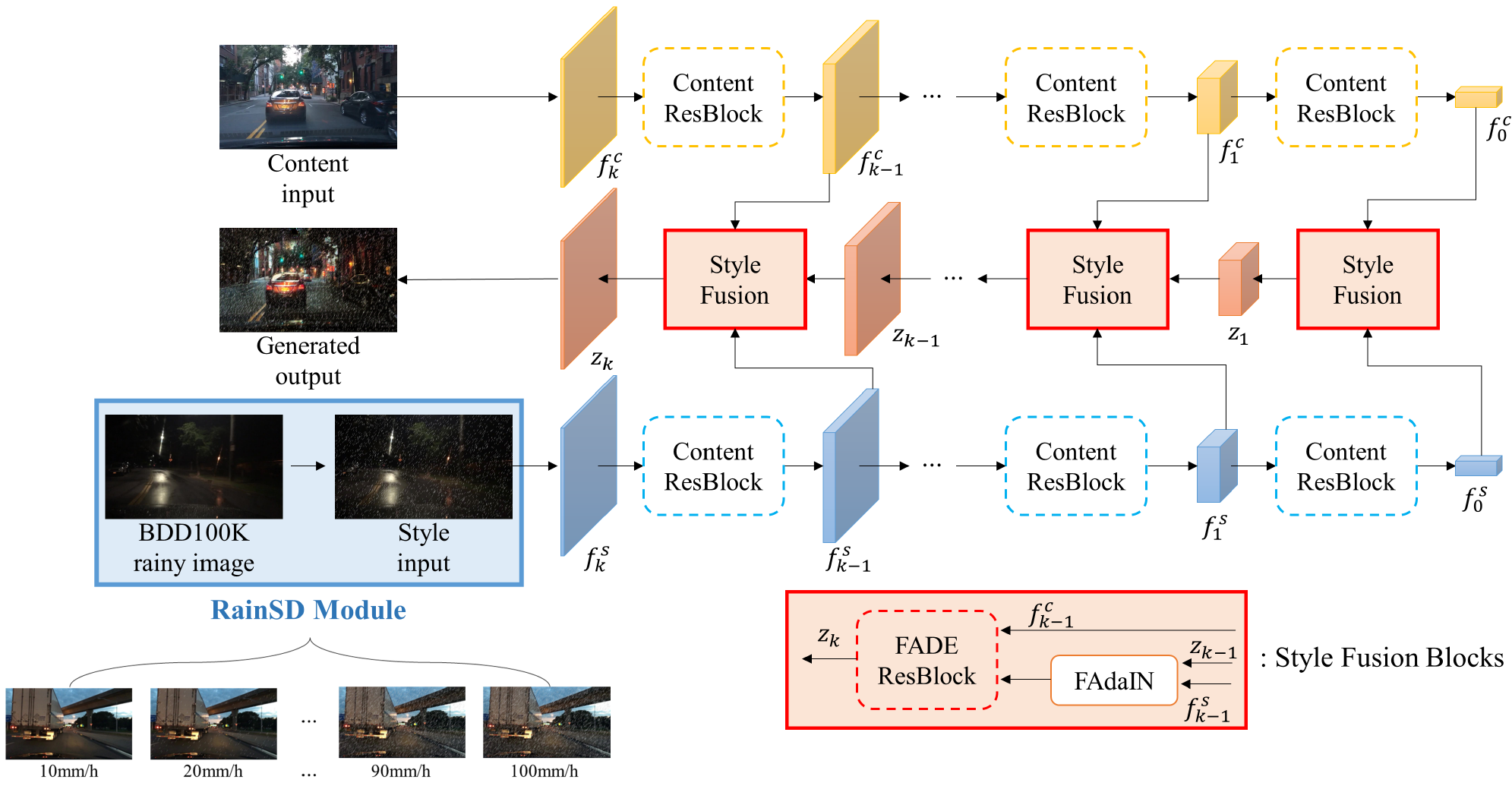}
\end{graphicalabstract}

\begin{highlights}
\item This study thoroughly analyses the degradation of image-based deep multi-task network for the autonomous vehicles under adverse weather conditions by heavy rain in the presence of diverse road conditions.
\item We propose our novel style diversification module for enhanced rain synthesis and establish a synthetic sensor blockage autonomous driving dataset generated from BDD100K datasets.
\item Based on the comparative evaluation results, we have discussed the impact of rain streaks generated through image-to-image translation process at the feature level on the performance of autonomous driving multi-task perception algorithms.
\end{highlights}


\end{frontmatter}

\section{Introduction}
\label{sec:introduction}

\begin{figure*}[h]     %
    \centering
    \includegraphics[width=1\linewidth]{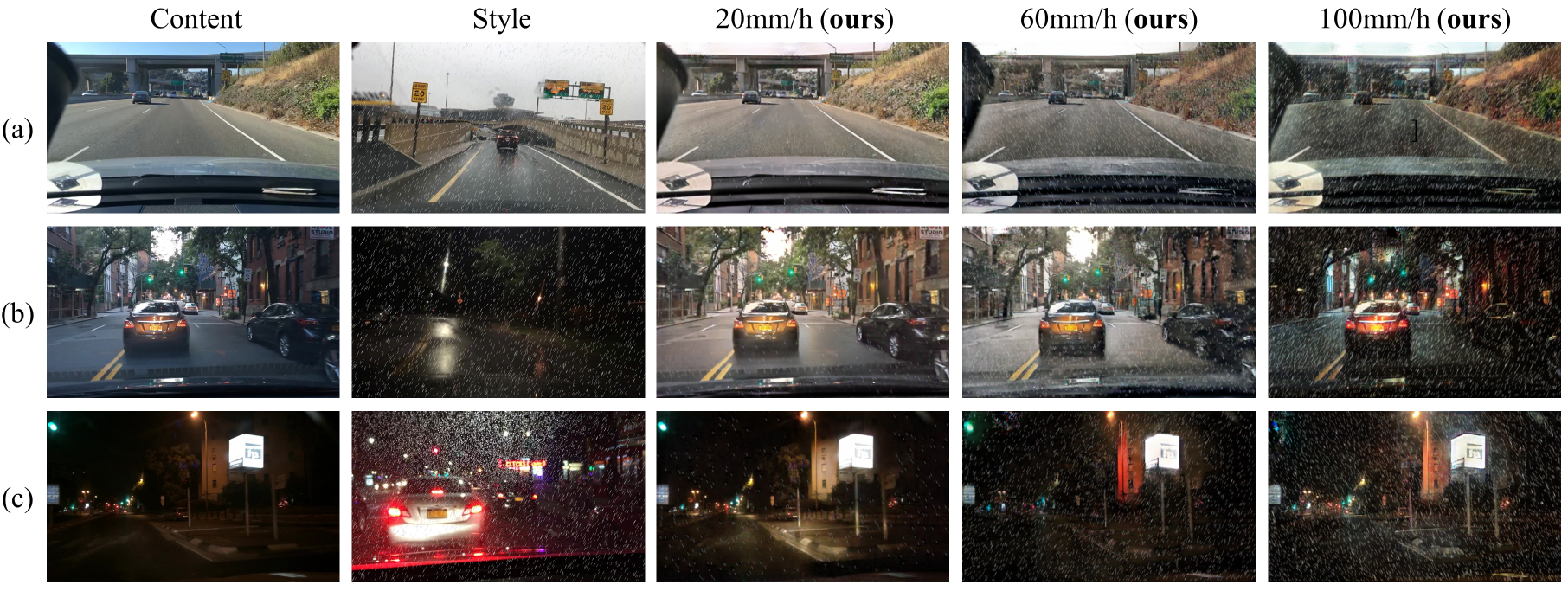}
\caption{The generated images fusing our rainy style diversification (RainSD) module and image-to-image translation model, including the rainfall rate of 20mm/h, 60mm/h, and 100m/h. (a) depicts translations from clear daytime to rainy daytime, (b) illustrates transitions from clear day to rainy night, and (c) displays clear night to rainy night transitions. Combined with feature-level style transfer based image-to-image translation model, our proposed method successfully generates fake rainy images with diverse rainfall rate preserving the existing semantic information of the content image.}
\label{Figure1}
\end{figure*}

Autonomous driving technology has been developed rapidly in recent decades, with increasing interests and numerous research studies supporting the application in various driving scenarios such as object detection, lane detection, lane keeping, autonomous emergency braking system, collision avoidance, and others \cite{1.hauer2019did}. More recently, many researchers and industries have been actively involved in the development of level 4 autonomous driving and are moving towards the development of level 5 technology \cite{2.ionita2017autonomous, 3.gkartzonikas2019have, 4.yeong2021sensor}. These levels of autonomous driving require the autonomous vehicles to successfully address various challenges from departure to arrival. Among many problems, one of the most frequent problems is the sensor blockage caused by the abnormal weather conditions \cite{5.jing2020determinants, 6.othman2021public, 7.cui2019review}. Especially, many applications such as lane detection and object detection have been developed based on deep neural networks (DNNs) which are highly influenced by the domain and the quality of datasets.

Among different applications, autonomous vehicles rely on core technologies such as lane detection, traffic object detection, and driving area segmentation, which play crucial roles in tasks like lane keeping, collision avoidance, and lane changing. Fig. 1 shows challenging situations involving the sensor blockage problem in autonomous vehicles under extremely heavy rain, generated using our image-to-image translation-based rain streak implementation. As depicted in Fig. 1, the autonomous vehicles operating in heavy rain condition face difficulties in capturing road scenes and detecting traffic objects necessary for algorithms based on the camera sensors. The damaged images obtained under adverse weather conditions are susceptible to mis-detection or failure of perception systems during autonomous driving. A severe accident could occur if there are other vehicles or pedestrian on the wrong trajectory generated by the mis-detected lane and object information.

Despite of the necessity of the research on the sensor blockage on the autonomous vehicles, the safety problem caused by sensor degradation under severe weather conditions has been treated roughly. Some researchers investigated the sensor blockage with only qualitative analysis \cite{8.hamzeh2020effect, 9.hnewa2020object, 10.goberville2020analysis} and discussed predictable outcomes without any quantitative results \cite{11.yoneda2019automated}. On the one hand, although the blockages can occur severe sensor failures that should be considered during risk assessment of autonomous vehicles, there are discussions on critical scenarios except for the sensor blockage caused by adverse weather conditions \cite{30.zio2018future}. This phenomenon on sensor blockage study happens because of the fact that different weather conditions are difficult to adjust, and it is very difficult to implement experiments under those conditions due to risk of accidents in real world environments \cite{9.hnewa2020object, 12.tang2020performance}. In the future, the developers in this relevant research field will need enormous amounts of qualified degraded image data to provide a feasible solution and discuss limitations of their methods during by developing novel neural networks. However, it is extremely hard to obtain those data in the real driving situations. One of the possible solutions is the usage of synthetic dataset that can be easily and repeatedly obtained from the virtual simulation environments \cite{13.espineira2021realistic}. Also, some studies have been conducted for data generation and data diversification for autonomous driving based on unsupervised image-to-image translation approaches \cite{31.hoffman2018cycada, 32.liu2017unsupervised, 33.zheng2020forkgan, 34.zhu2017toward, 35.xia2023image} to overcome night-to-day domain shift \cite{36.schutera2020night, 37.arruda2019cross, 38.lin2020gan} and rainy road conditions \cite{39.jin2021raidar, 40.jeong2021memory}.

Regarding the difficulties of the sensor blockage studies on the autonomous vehicles, our goal in this paper is to generate a useful synthetic weather condition dataset with rainy driving scenes and to evaluate the performance of the multi-task autonomous perception algorithms under those scenes. Thus, we propose a novel dataset generated from BDD100K \cite{41.yu2020bdd100k} based on our rainy style variation module combined with image-to-image translation method named TSIT \cite{42.jiang2020tsit} and the evaluation format for multi-task perception algorithms including traffic object detection, lane detection, and driving area segmentation. Utilizing our novel dataset that encompasses diverse weather conditions, including both clear and rainy scenes, a multi-task network has been trained and executed. Evaluation of the network performance has been conducted using the BDD100K dataset \cite{41.yu2020bdd100k}, which features real road data captured under clear and rainy weather conditions. The original contributions of this paper are as follows:

\begin{itemize}
\item This study thoroughly analyses the degradation of image-based deep multi-task network for the autonomous vehicles under adverse weather conditions by heavy rain in the presence of diverse road conditions (i.e., normal condition, rainy condition with light rain, and rainy condition with extremely heavy rain), both qualitatively and quantitatively.
\item To evaluate the multi-task perception networks based on deep learning, we propose our novel style diversification module for enhanced rain synthesis and establish a synthetic sensor blockage autonomous driving dataset generated from BDD100K datasets. We have also provided the labelled train and test set with various rainfall rates in the range from 0mm/h to 100mm/h.
\item Based on the comparative evaluation results, we discuss the impact of rain streaks generated through image-to-image translation process at the feature level on the performance of autonomous driving multi-task perception algorithms.

\end{itemize}

\section{Related Works}
\label{sec:RELATEDWORKS}

\begin{figure*}[h]     %
    \centering
    \includegraphics[width=1\linewidth]{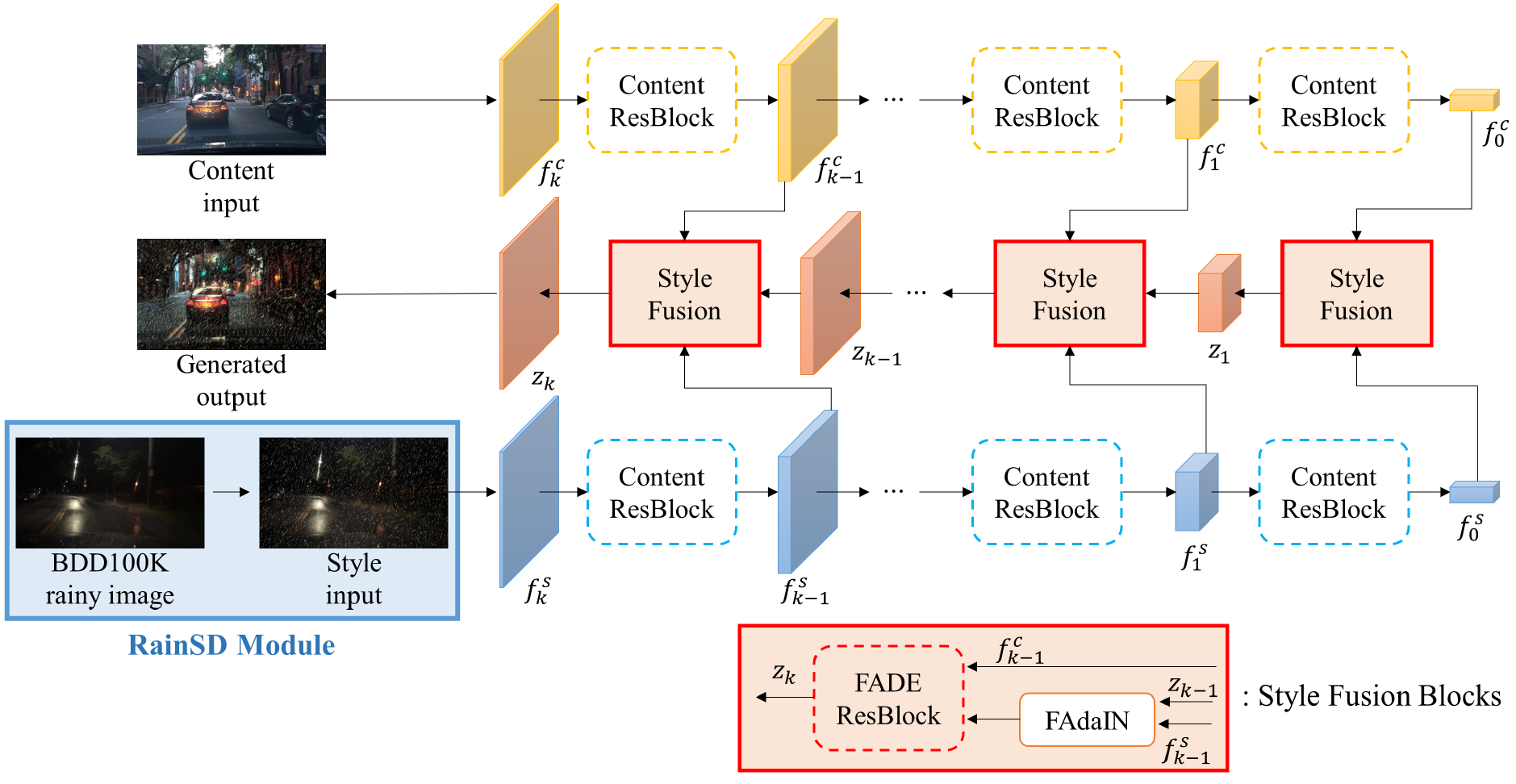}
\caption{The overall structure of the proposed network with our novel RainSD module. Style fusion blocks include FAdaIN style transfer module and FADE ResBlock from \cite{42.jiang2020tsit}. The network is consisted of two streams of encoder-decoder structure, which enables the network to learn style distribution of both forward feature and backward feature. A generation process of the synthetic image is highly inspired from \cite{42.jiang2020tsit}, and we attempt to vary the styles of the rainy scenes, especially by generating realistic rain streaks based on application of the experimentally obtained equation. Detailed explanation on RainSD module is in Fig.4}
\label{Figure2}
\end{figure*}

\subsection{Autonomous Driving Datasets}
Autonomous driving datasets contain rich data necessary for vehicles to understand and manipulate their surrounding environments. These datasets primarily encompass information obtained from various sensors such as cameras, Radar, LiDAR, and others, capturing diverse scenarios involving road conditions, vehicles, pedestrians, and other surrounding objects. Largely, we can categorize autonomous driving datasets into two categories, real road dataset and synthetic framework dataset.

There are numerous road datasets accessible for the training of autonomous driving tasks, such as Cityscapes \cite{43.cordts2016cityscapes}, BDD100K \cite{41.yu2020bdd100k}, KITTI \cite{44.geiger2013vision}, and nuScenes \cite{45.caesar2020nuscenes}. Those datasets provide real roads scenarios with various weather conditions and time domains. They have significantly contributed to the development of deep learning-based perception technologies, enabling vehicles to comprehend and make real-time judgments during time varying driving conditions. Advanced deep learning architectures such as Convolutional Neural Networks (CNNs), Recurrent Neural Networks (RNNs), and Transformers process visual and sensor data, providing a sophisticated understanding of the driving environment, so that the autonomous vehicles nowadays exhibit an enhanced capability to interpret driving scenarios in a higher-dimensional manner.

Some evaluation environments for autonomous driving have been developed by focusing on the virtual sensor data. SYNTHIA \cite{14.ros2016synthia} and Synscapes \cite{15.wrenninge2018synscapes} are popular synthetic datasets used for training and evaluation of semantic seg-mentation, domain adaptation, other applications on autonomous driving. Recently, there are several platforms for testing all kinds of autonomous vehicle’s modules (i.e., perception, planning and control). There are some open-source simulators such as CARLA \cite{16.dosovitskiy2017carla}, LGSVL \cite{17.rong2020lgsvl}, and AirSim \cite{18.shah2018airsim}. These provide not only the visual data such as RGB images, depth images, and point clouds, but also the implementation of the planning and control algorithms. Also, they are user-friendly programs providing Python API which is easy to control the intrinsic parameters. Those features can boost the research on autonomous vehicles with their convenient connectivity of three modules for driving.

There are several studies attempting to facilitate autonomous driving research under rainy conditions. To generate the sensor blockage similar to the real rainy scenes, \citet{19.jeon2022carla} developed a metric-based similarity comparison and image generative method . They compared the real rainy images and the synthetic rainy images to prove the similarity and established the experimental rainfall equation. Upon the equation, the performance of traditional lane detection method was evaluated through steering angle calculated by pure pursuit algorithm. \citet{39.jin2021raidar} proposed a newly annotated image dataset named RaidaR, which contains the largest size of rainy road scenes. They have generated the images through adding some masks upon unpaired image-to-image translation algorithms. The dataset was undergone through a thorough evaluation involving state-of-the-art semantic segmentation networks \cite{48.zhu2019improving, 49.tao2020hierarchical} and instance segmentation networks \cite{50.kirillov2020pointrend, 51.chen2019tensormask}.

\subsection{Data Generation via Image-to-Image Translation}
So far, image-to-image translation has been utilized across various domains \cite{47.pang2021image}, including its application in the development of datasets for autonomous driving on roads. Different companies and research institutions require diverse simulations and real-world data to train and test autonomous vehicles. Image-to-image translation is used to increase the quantity of data and enhance its diversity \cite{53.lee2018diverse}. By transforming data acquired from various road environments and creating new scenarios, this technology enriches training datasets for autonomous vehicles. It can generate diverse road scenarios in a given virtual environment, making them similar to real-world road conditions \cite{54.pahk2023effects}. Virtual road data includes various weather conditions, traffic situations, and road signs, contributing to the improvement of autonomous driving algorithms. Image-to-image translation contributes to improving the robustness of algorithms in given road situations. This technology is instrumental in training models to adapt to changes in diverse environments and perform stable autonomous driving \cite{52.pizzati2020domain}. 

Pix2Pix \cite{55.isola2017image} is a model that employs conditional Generative Adversarial Networks (GANs) to learn mappings between input images and their corresponding output images. This model is utilized for generating specific output images for given input images. For example, it can transform black-and-white sketches into real photographs. CycleGAN \cite{56.zhu2017unpaired} is a GAN-based model utilized for the transformation of images between two domains. For instance, it can be employed to convert images of horses into those of zebras or to transform summer landscapes into winter scenes. This model is trained by emphasizing cycle consistency in image translation tasks. 

MUNIT \cite{57.huang2018multimodal} is a model designed for performing multimodal image translation, enabling flexible mapping of style and content between input and output images. This capability allows for unrestrained image transformations between various domains. DRIT++\cite{58.lee2020drit++} tackles the challenges inherent in image-to-image translation, specifically addressing the issues of lacking aligned training pairs and the uncertainty stemming from multiple potential outputs for a single input image. The method employs a disentangled representation, incorporating the embedding of images into both a domain-invariant content space and a domain-specific attribute space. TSIT \cite{42.jiang2020tsit} is a straightforward and versatile image-to-image translation framework that emphasizes the significance of normalization layers. It utilizes a carefully crafted two-stream generative model incorporating novel coarse-to-fine feature transformations, allowing for the effective capture and fusion of multi-scale semantic structure information and style representation.

\subsection{Multi-task Learning for Autonomous Driving}
Multi-task learning is utilized in autonomous driving systems to perform various tasks simultaneously \cite{60.chen2018multi}. For instance, it can handle tasks such as object detection, lane-keeping, and driving area segmentation concurrently in autonomous vehicles. By learning multiple tasks, it efficiently utilizes data, and the knowledge gained from one task can enhance performance in other tasks, contributing to increased data efficiency and improved model generalization \cite{61.zhang2021survey}. Also, multi-task learning enables the end-to-end training of multiple tasks \cite{62.liu2019end}, simplifying the design and training of complex autonomous driving systems. Additionally, the model's ability to learn from multiple tasks enhances its adaptability to diverse scenarios.

YOLOP \cite{63.wu2022yolop} is a panoptic driving perception network designed for simultaneous traffic object detection, driving area segmentation, and lane detection. The model, featuring one encoder for feature extraction and three decoders for task-specific handling, demonstrates outstanding performance on the challenging BDD100K dataset. \citet{64.vu2022hybridnets} developed an end-to-end perception network, HybridNets, for simultaneous multi-tasking in autonomous driving. The study proposes optimizations such as an efficient segmentation head, customized anchors, and a training loss function to enhance accuracy. The authors propose efficient segmentation and object detection heads utilizing a weighted bidirectional feature network, introducing a real-time drivable multi-task network designed for autonomous driving.

\section{Methodology}
\label{sec:methodology}

\begin{figure*}[h!t]     %
    \centering
    \includegraphics[width=1\linewidth]{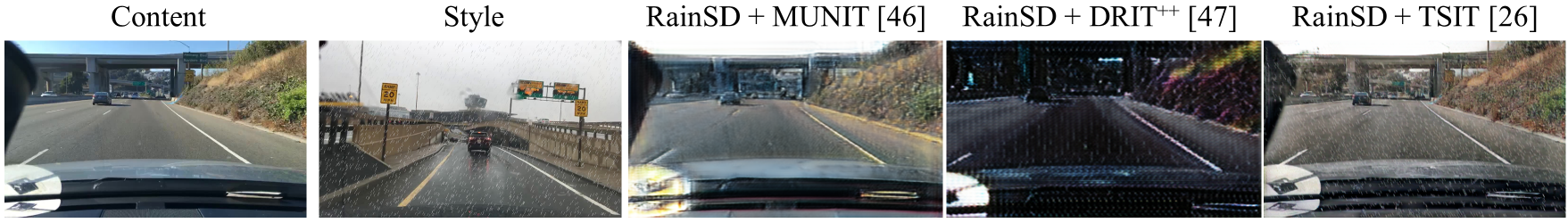}
\caption{Comparison of the generated images from the representative image-to-image translation models, comprised of MUNIT, DRIT++, and TSIT. Both semantic information and the style information are rarely transferred when using MUNIT and DRIT++, making unintended distortion to the crucial objects in the road scenes. TSIT successfully translates the style information from the rainy images possess to the content image, preserving the important semantic structures.}
\label{Figure3}
\end{figure*}

\subsection{Data Generation via Image-to-Image Translation}
\label{sec:DataGeneration}

In the context of image-to-image translation for data generation, a multitude of approaches have been investigated to produce diverse and lifelike images without requiring explicit labeling. Prominent models within this domain encompass MUNIT \cite{57.huang2018multimodal}, DRIT++ \cite{58.lee2020drit++}, and TSIT \cite{42.jiang2020tsit}. These methodologies share a salient advantage by operating in an unsupervised fashion, depending solely on paired images to accomplish the intended style transfer. In this study, we generate our rain synthesis dataset generated through TSIT, using clear weather images from BDD100K dataset \cite{41.yu2020bdd100k}. Fig.2 shows the overall workflow of this study. The image-to-image translation model needs content images and style images, so that the style information of the style image is transferred and fused to the semantic structures of the content image. BDD100K dataset has explanations on some specific properties such as weather, location of the scene, and time of day. Among them, we used the frame attributes regarding the weather and extracted clear images and rainy images from the original BDD100K dataset. Afterwards, we made the content image set with the original clear images and the style image set by implementing our rainy style diversification (RainSD) module on the original rainy images. 
 
TSIT stands out as an innovative approach that incorporates both semantic structure information and style representation during the image-to-image translation process. By adopting a two-stream architecture, TSIT effectively captures and fuses multi-scale features related to content and style. This dual-stream mechanism enables a smoother preservation of content, making TSIT particularly advantageous in scenarios where maintaining the integrity of road object details is crucial. Conventional image-to-image translation models, such as MUNIT and DRIT++, are powerful in generating diverse images, but often face challenges in distorting crucial road objects such as lanes, vehicles, or sidewalks. The inherent drawback lies in their generic approach, which may not prioritize the preservation of critical semantic information. This limitation raises concerns about the reliability of generated images for applications of autonomous driving, where accurate representation of road scenes is necessary. Fig.3 shows the generated images from three image-to-image translation models, including MUNIT, DRIT++, and TSIT. As explained in Fig.3, the road scenes of MUNIT and DRIT++ are severely ruined, whereas TSIT generates fake images of which semantic structures are preserved well. Also, the style information like cloudy atmosphere and the rain streaks scattered in the air.

\begin{figure}[h!b]     %
    \centering
    \includegraphics[width=1\linewidth]{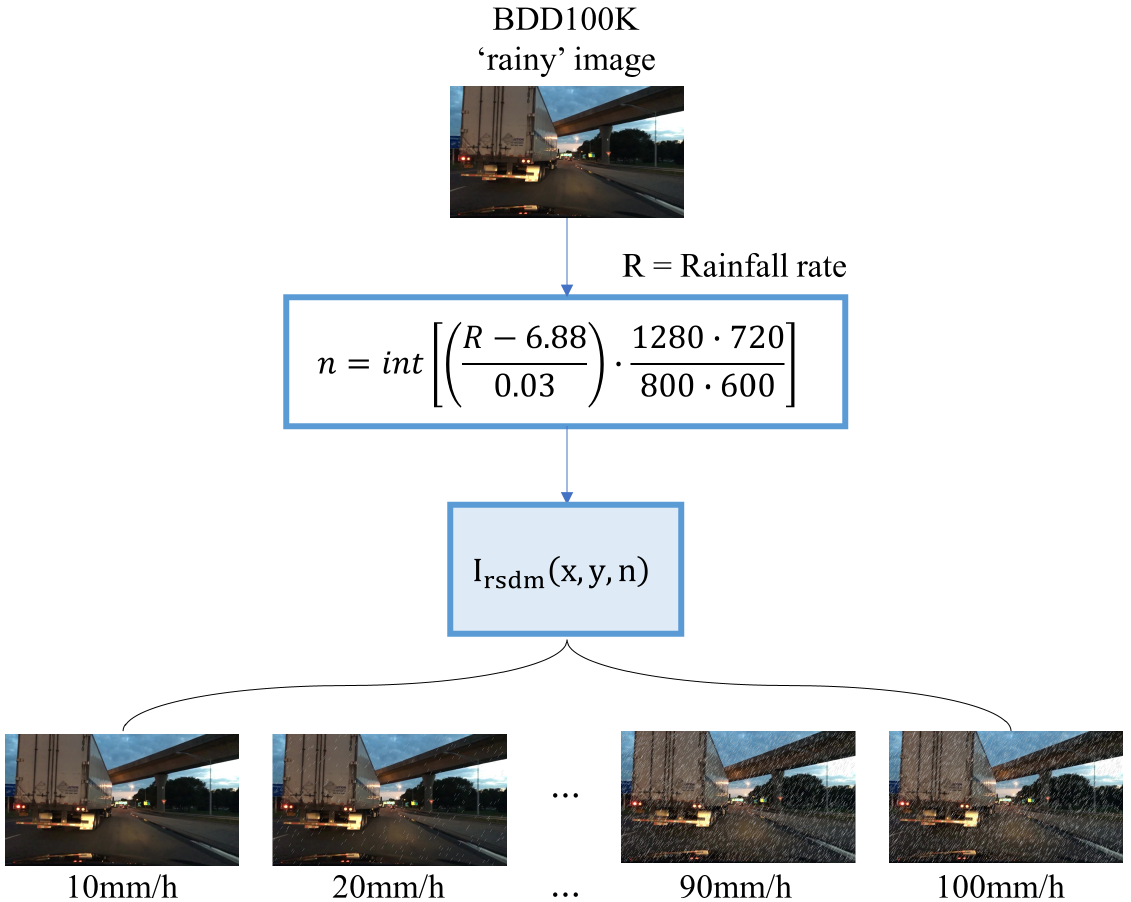}
\caption{The structure of the proposed RainSD module.}
\label{Figure4}
\end{figure}

\begin{figure*}[h!t]    
    \centering
    \includegraphics[width=1\linewidth]{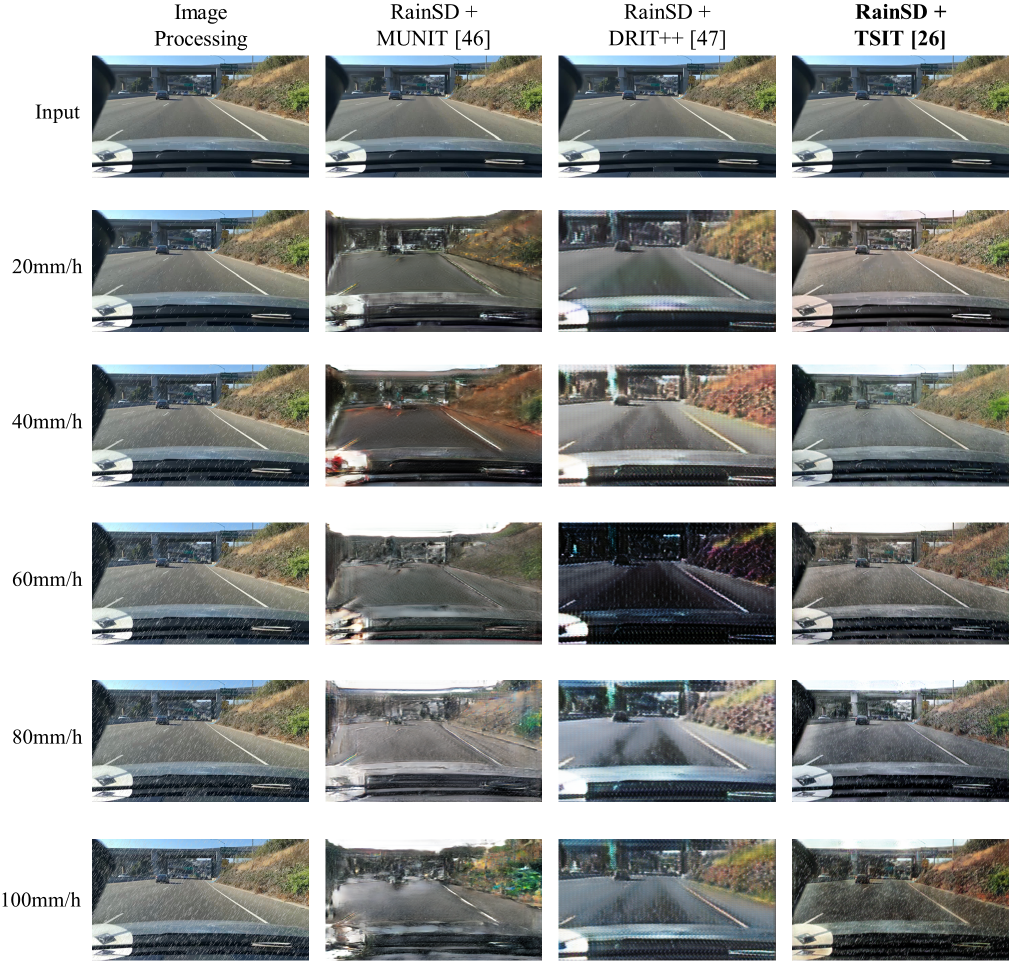}
\caption{Qualitative results of our novel RainSD module combined with different kinds of image-to-image translation methods. It shows the images generated from clear-daytime (test A, content image) to rainy-daytime (test B, style image). The images depict the results of image synthesis process with regard to the rainfall amounts including 20mm/h, 40mm/h, 60mm/h, 80mm/h, and 100mm/h, upon consideration of realistic calculation \cite{19.jeon2022carla} using the method proved by image similarity metrics. MUNIT and DRIT++ fail to deliver the amount of rain streaks in each style image, whereas TSIT successfully conducts style transfer.}
\label{fig:Figure5}
\end{figure*}

\subsection{Rainy Style Diversification (RainSD) Module}

To advance research in autonomous driving, datasets such as BDD100K and RaidaR \cite{39.jin2021raidar} have been made publicly available. However, existing studies have primarily focused on capturing road styles like wet roads and droplets on the window shields in scenarios with minimal rainfall or using image generation models to transform clear weather images into rainy styles. When using such datasets, it becomes challenging to investigate the limitations of autonomous vehicles in heavy rainfall conditions compared to what the models have been trained on or to develop systems to overcome these challenges. To address this gap, we propose a style diversification module. This module aims to leverage the authentic rainy styles present in the existing BDD100K dataset while also introducing artifacts such as rain streaks that may occur in more intense rainfall situations. This approach allows for a more comprehensive exploration of the limitations posed by heavy rain on autonomous vehicles and the development of systems to address these challenges. Fig. 4 shows the structure of RainSD module.

To infuse artifacts into the original rainy images of the BDD100K dataset, rain streak layers are strategically placed. Eq.(1) illustrates the process of transforming the input image into a simulated rainy scene.

\begin{equation}
I_{\text{rsdm}}(x,y) = \sum\limits_{i=1}^n Line(x-\text{rand(interval)}, y)
\end{equation}
where $I_{rsdm}$ represents the image generated through RainSD module, with $(x,y)$ denoting coordinates within the image's width and height range. $Line$ corresponds to the formation of each rain streak, where the $interval$ determines the spacing between streaks, and the streaks are randomly positioned. N represents the rainfall amount, influencing the creation of varying numbers of rain streaks at random locations within the image. The formula for determining the number of rain streaks based on rainfall amount is derived from experimental equation from \cite{19.jeon2022carla}. Given the resolution of images from BDD100K is 1280x720, exceeding 800x600 of the original equation, adjustments of the number of the rain streaks are made to accommodate the higher resolution. Then, the images undergo the TSIT \cite{42.jiang2020tsit} process to transform them into realistic rainy scenes. The style information for rainy images is derived from  feature adaptive instance normalization (FAdaIN) style transfer module during the training process, introduced from \cite{42.jiang2020tsit}. Eq.(2) represents the formulation of FAdaIN used in our training.
\begin{equation}
\vspace{-0.2cm}
FAdaIN(z_{i},f_{i}^{\text{rsdm}}) = \sigma(f_{i}^{\text{rsdm}}) \frac{z_{i}-\mu(z_{i})}{\sigma(z_{i})}+\mu(f_{i}^{\text{rsdm}})
\end{equation}
where $z_{i}$ denotes the feature map after $i$-th FADE residual block in TSIT generator, and $f_{i}^{rsdm}$ corresponds to $i$-th feature-level style distribution from the style stream, which contributes to the diversity of rainfall amounts in the generated image.

\begin{figure*}[h!t]    
    \centering
    \includegraphics[width=1\linewidth]{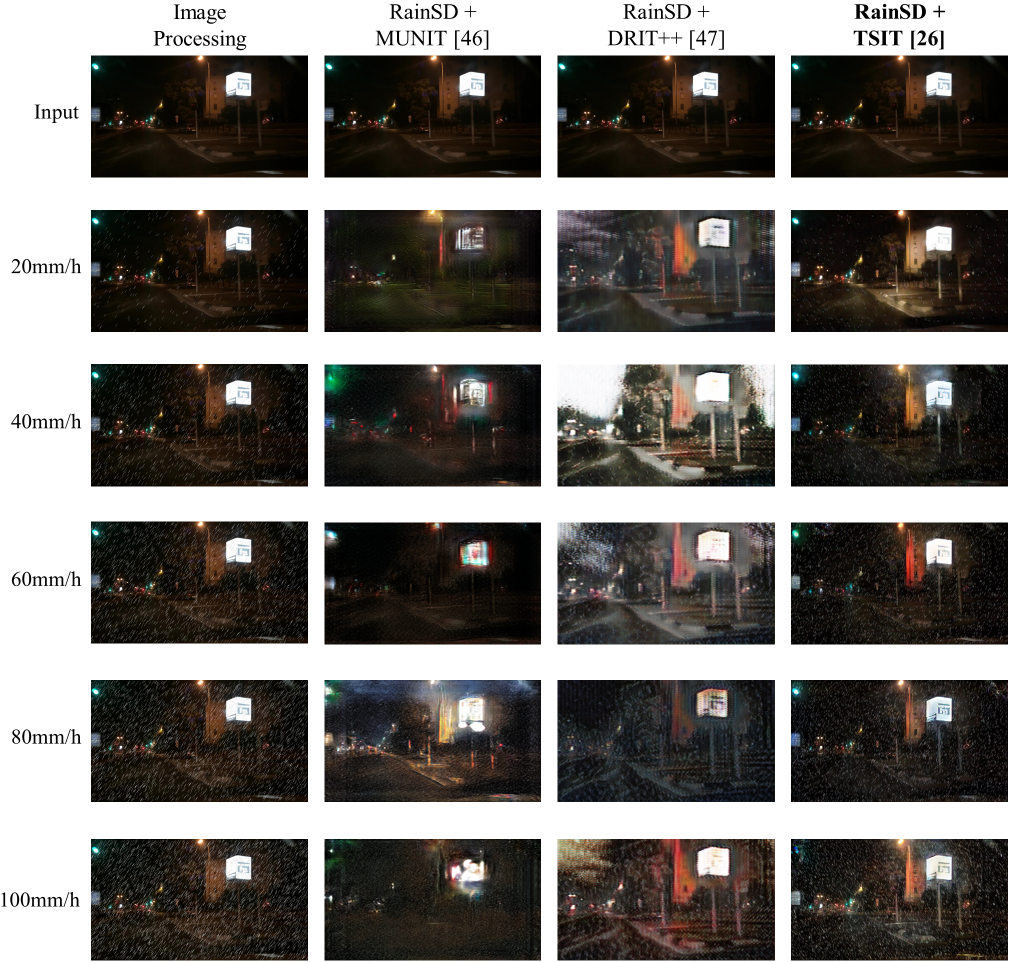}
\vspace{-0.2cm}
\caption{Qualitative results of our novel RainSD module combined with image-to-image translation methods. It shows the images generated from clear-night (test A, content image) to rainy-night (test B, style image). DRIT++ severely ruins the style of the given image, making the daytime style at 40mm/h rainfall rate images. Also, MUNIT fails to convert the rainfall amount of the given style image at 20mm/h, 40mm/h, and 60mm/h. TSIT generates the appropriate number of rain lines in the image, and makes the rain streaks more realistic than those from image processing method.}
\label{fig:Figure6}
\end{figure*}
\vspace{-0.2cm}

\begin{equation}
L_{G} = -\mathbb{E}[D(g)]+\lambda_{P}\mathcal{L}_{P}(g,x^{c})+\lambda_{FM}\mathcal{L}_{FM}(g,x^{s}),
\end{equation}
\begin{equation}
L_{D} = -\mathbb{E}[min(-1+D(x^{s}),0]-\mathbb{E}[min(-1-D(g),0],
\end{equation}
where $g = G(z_{0},x^{c},x^{s})$ represents the generated image, and $z_{0}$, $x^{c}$, $x^{s}$ are the input noise map in the latent space, the content image, and the style image, respectively. $\mathcal{L}_{P}$ and $\lambda_{P}$ are the perceptual loss \cite{65.johnson2016perceptual} minimizing the gap between the feature representations from VGG-19 backbone and the corresponding weight. $\mathcal{L}_{FM}$ and $\lambda_{FM}$ \cite{66.wang2018high} are the feature matching loss that matches different features from each layer of the discriminator and the corresponding weight.

\vspace{-0.3cm}
\section{Experiments}
\label{sec:Experiment}

\begin{figure*}[h!b]    
    \centering
    \includegraphics[width=13cm]{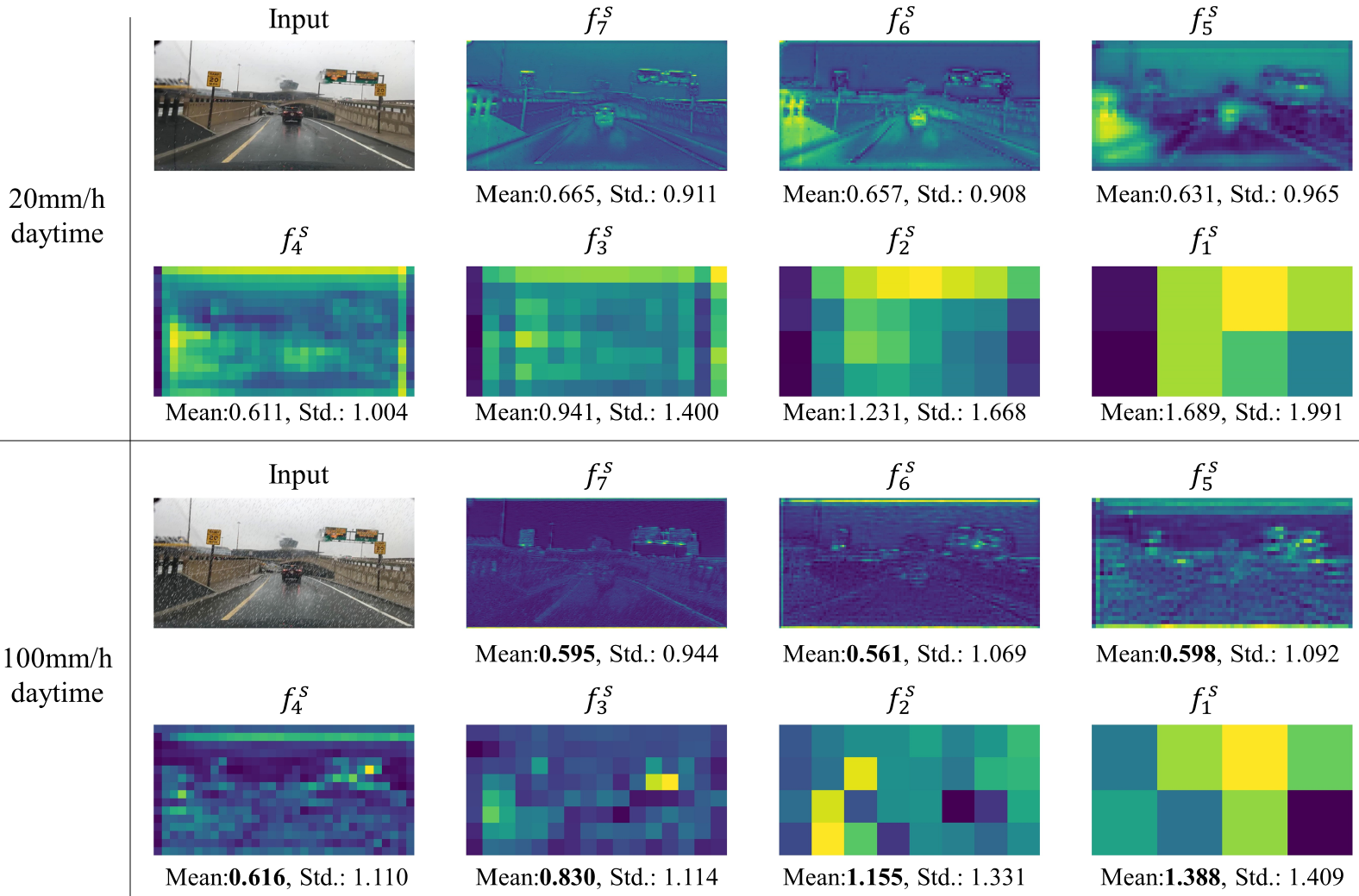}
\caption{Feature maps of the daytime style image and the following style distribution information per channel from each layer of the ResBlock-based TSIT generato. As $k$ becomes larger and the layers get deeper, the mean and the standard deviation values start to converge, which means that the feature gradually deliver the style information of the style image. In the daytime, the values in the feature map become much smaller than 20mm/h when the rainfall rate is 100mm/h.}
\label{fig:Figure7}
\end{figure*}

\subsection{Objectives}
The objective of the TSIT in this work to maximize the generator loss Lg and minimize the discriminator loss Ld. The generator loss and the discriminator loss are as follows:
\vspace{-0.2cm}

\vspace{-0.1cm}
\subsection{Datasets}

\newcolumntype{V}{>{\centering}X}%

\begin{table*}[h!t]
	\small
	\caption{Quantitative results from the state-of-the-art multi-task networks on BDD100K dataset. The results in bold red letters indicate the highest performance, while those in bold blue letters signify the lowest performance, in each task of two respective networks. BDD represents image set from the original BDD100K dataset, \textbf{RainSD} designates the original BDD100K image set combined with the rain made by only image processing-based RainSD module, and $\textbf{RainSD}+\textbf{TSIT}$ means the dataset generated by TSIT integrated with our RainSD module, which generates rain streaks through the forward/backward feature-level style transfer.}
	\label{tab:Results} 
\begin{tabular}{@{}cccccccc@{}}
\toprule
                             &                                                                                         & \multicolumn{2}{c}{Traffic object detection}                                & \multicolumn{2}{c}{Driving Area Seg.}                               & \multicolumn{2}{c}{Lane Detection}                                          \\ \cmidrule(l){3-8} 
\multirow{-2.5}{*}{Network}    & \multirow{-2.5}{*}{\begin{tabular}[c]{@{}c@{}}Training Data\\ (clear:rainy=4070:4070)\end{tabular}} & Recall (\%)                               & mAP (\%)                         & mIOU (\%)                            & Acc. (\%)                        & mIOU (\%)                            & Acc. (\%)                        \\ \midrule
                             & BDD:BDD                                                                                 & 87.2                                 & {\color[HTML]{FE0000} \textbf{10.9}} & {\color[HTML]{FE0000} \textbf{75.7}} & {\color[HTML]{FE0000} \textbf{86.2}} & {\color[HTML]{FE0000} \textbf{22.0}} & {\color[HTML]{FE0000} \textbf{68.9}} \\
                             & BDD:BDD+\textbf{RainSD}                                                                          & {\color[HTML]{3531FF} 84.2}          & {\color[HTML]{3531FF} 4.39}          & {\color[HTML]{3531FF} 69.4}          & 84.8                                 & {\color[HTML]{3531FF} 17.1}          & {\color[HTML]{3531FF} 65.4}          \\
\multirow{-3}{*}{HybridNets \cite{64.vu2022hybridnets}} & BDD:BDD+\textbf{RainSD}+\textbf{TSIT}                                                                     & {\color[HTML]{FE0000} \textbf{87.3}} & 5.84                                 & 72.9                                 & {\color[HTML]{3531FF} 84.5}          & 20.8                                 & 68.6                                 \\ \midrule
                             & BDD:BDD                                                                                 & 65.9                                 & {\color[HTML]{3531FF} 10.1}          & {\color[HTML]{FE0000} \textbf{86.9}} & {\color[HTML]{FE0000} \textbf{95.9}} & {\color[HTML]{3531FF} 34.3}          & {\color[HTML]{3531FF} 39.5}          \\
                             & BDD:BDD+\textbf{RainSD}                                                                          & {\color[HTML]{3531FF} 64.8}          & {\color[HTML]{3531FF} 10.1}          & {\color[HTML]{3531FF} 84.1}          & {\color[HTML]{3531FF} 94.9}          & {\color[HTML]{FE0000} \textbf{36.4}} & {\color[HTML]{FE0000} \textbf{44.2}} \\
\multirow{-3}{*}{YOLOP \cite{63.wu2022yolop}}      & BDD:BDD+\textbf{RainSD}+\textbf{TSIT}                                                                              & {\color[HTML]{FE0000} \textbf{66.1}} & {\color[HTML]{FE0000} \textbf{11.7}} & 85.2                                 & 95.3                                 & 35.6                                 & 42.2                                 \\ \bottomrule
\end{tabular}
\end{table*}

We perform dataset expansion and augmentation for multi-task evaluation. Firstly, we extract a total of 5070 rainy BDD100K images to create the training set for the TSIT network. Then, using TSIT fused with RainSD module, we generate images. Applying the RainSD module for rainfall rates ranging from 10mm/h to 100mm/h in 10mm/h intervals, we generate 50700 images for the trainB set, which are style images. Additionally, we extract the same number of clear BDD100K images to create the trainA image set, serving as content images. Once TSIT training is completed, we extract clear BDD100K images not used in training to create the testA set with 4070 images. Consequently, our dataset comprises 4070 clear images and 4070 novel rainy images obtained through TSIT combined with the RainSD module, providing a 1:1 representation of road scenarios in clear and rainy weather conditions. Using these datasets, we evaluate multi-task networks for autonomous vehicles. 

\subsection{Implementation Details}
The proposed RainSD module is an additive system that can be applied on the existing image-to-image translation methods. In this work, TSIT underwent training in the MMIS (multimodal image synthesis) task configuration for image-to-image translation, employing a batch size of 1. The model was optimized using the Adam optimizer with an initial learning rate set at 0.0002. The training spanned 20 epochs without any decay in the learning rate. Weight parameters were assigned with $\lambda_{P}=1$ and $\lambda_{FM}=1$. 
All experiments were designed using the PyTorch framework \cite{67.paszke2019pytorch}. The image-to-image translation model was trained using a single NVIDIA RTX 4090 GPU, while the multi-task networks were tested using a NVIDIA GTX TITAN XP, trained under the same conditions.

\begin{figure*}[h!t]    
    \centering
    \includegraphics[width=13cm]{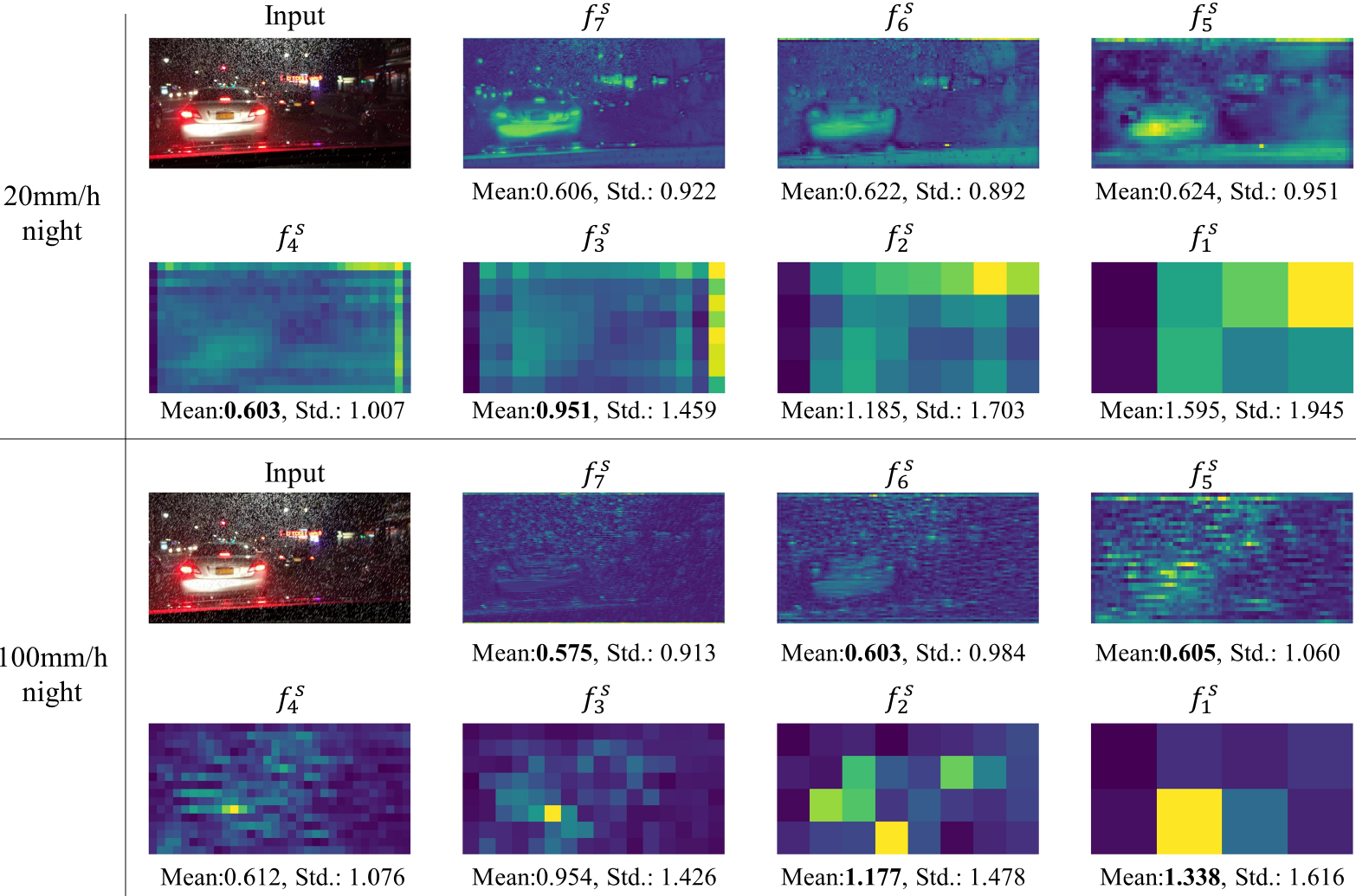}
\caption{Feature maps of the night time style image and the following style distribution per channel from each layer of the ResBlock-based TSIT generator. Values in bold letter are the smaller ones.  Similar to the daytime, for deep features, the mean and the standard deviation become smaller, which means that the feature gradually deliver from the semantic information the content image to the style information of the style image. Different from the daytime images, the differences of values in the feature map are not large.}
\label{fig:Figure8}
\end{figure*}

\subsection{Experimental Results}
\label{sec:Experiment}
For data generation experiment, Fig.5 and Fig.6 illustrate the images generated for different rainfall rates at daytime and night, respectively. Fig.5 provides visualization results of style transfer from clear day images to rainy day images for different rainfall amounts. Our RainSD module is applied to various image-to-image translation models, revealing performance variations among the models. In the case of MUNIT, although the road surface style is well transferred, rain streaks are barely translated, and semantic information of crucial objects such as lanes and vehicles in the forward view is significantly distorted, particularly in models trained with high rainfall amounts. For DRIT++, rain streaks are somewhat transferred, but the road style information is severely disrupted, resulting in a style transfer that differs from reality. In contrast, TSIT successfully transfers rainfall amounts as intended by the RainSD module, generating images where both content and style information of lane and road details are well preserved without distortion. Figure 6 displays results of the image generation from clear night images to rainy night images for different rainfall amounts. Similarly, there are performance differences among image generation models. For MUNIT, rain streaks are either not transferred according to rainfall amounts, or when transferred, the results deviate significantly from the intended style. Regarding DRIT++, both road style information and content information are severely disrupted, and the style transfer differs completely from the intended one. Lastly, TSIT successfully transfers rainfall amounts as the RainSD module intended, preserving both content and style information of lane and road details without distortion, while appropriately generating rain streaks according to the rainfall amount.

For multi-task learning evaluation, Table \ref{tab:Results} shows the quantitative results of the experiment on the state-of-the-art multi-task networks widely implemented on BDD100K dataset. We used HybridNets and YOLOP for the performance evaluation of the task networks. 

Overall, in both HybridNets and YOLOP results, the lowest performance was observed when training with the RainSD-only applied train set. HybridNets exhibited the best performance when using the original BDD100K set as the training dataset. While the performance was not consistently the best when applying RainSD and TSIT together, it generally showed better performance compared to methods without TSIT. Evaluation against the original BDD100K training set revealed higher performance in object recognition recall by 3.7\%, drivable area segmentation mIOU by 5.0\%, lane recognition mIOU by 21.6\%, and accuracy by 4.9\%. Consequently, in the face of severe rainfall, the performance of deep learning networks may decline, but the proposed module-based learning approach in HybridNets proved more effective for environmental perception than learning rain images through traditional computer vision techniques.

Next, in YOLOP, even when using RainSD-only set, it showed better performance than training with the original BDD100K data, with a 1.8\% improvement in object recognition recall, 15.8\% in mAP@0.5, 3.8\% in lane recognition mIOU, and 6.8\% in accuracy. While the original BDD100K training set exhibited the best performance in driving area segmentation, YOLOP also showed high performance in this task across the other training settings using RainSD module. Therefore, if using the YOLOP model for autonomous driving, training the vehicle with RainSD-applied images generated through TSIT can lead to better driving performance of the vehicle.

\section{Discussion}

Analyzing the results from Fig.5 and Fig.6, a noticeable trend was identified. To investigate the style transfer process occurring during the transformation of these images, we extracted style distributions at the feature level. As observed in Fig.5, there is a tendency for the output images generated by training with increasingly heavy rainfall to become darker. Particularly interesting is the fact that the style image used in this setting was also an image captured during daytime conditions, which made us suggest a significant influence due to heavy rainfall. In the case of Fig.6, where there were no significant changes in the images as in the daytime scenario, style distributions were also extracted for tracing both distributions from the images of daytime and night time. The feature maps of the style images used for each image generation and the corresponding style distributions are represented in Fig.7 and Fig.8.

As depicted in Fig.7, we can observe significant differences in the mean and standard deviation values representing the style information of features closer to the input image level, such as f7 or f6. In particular, regarding the mean value, f7 is 10.5\% lower, and f6 is 14.6\% lower compared to the feature of the 20mm/h style image when the rainfall is 100mm/h. This indicates that the average values of the features are distributed lower at each position, suggesting a higher likelihood of forming lower brightness as the style image experiences heavier rainfall. This phenomenon can be interpreted as the result of dense clouds often present in rainy and overcast weather causing lower illumination, or it may be influenced by the color of rain streaks artificially altering the distribution of style. In the former case, experiencing reduced visibility due to heavy clouds on rainy days aligns with the lower brightness that can be observed in the real world. In the latter case, it represents a characteristic that can be effectively utilized for future image generation studies. Through this work, although establishing a precise correlation between the degree of cloudiness and rainfall intensity on rainy days seems challenging, we discovered a plausible association, providing a potential factor for consideration in future development of rainy image generation models.

\section{Conclusion}
In this work, we present our style diversification method for enhanced rainy image synthesis, referred to as RainSD. Our primary focus is on the generation of rain streaks that can appear under heavy rain conditions while autonomous driving. Through experiments in generating the synthesized BDD100K dataset, we demonstrate the advantages of the proposed RainSD module fusing with existing style transfer-based image-to-image translation method, TSIT. Among various image-to-image translation approaches, we found that the combination of feature-level style transfer and our proposed module works well. We also evaluated our generated dataset on the state-of-the-art autonomous drivings based on multi-task learning. According to the performance evaluation results, HybridNets better fits with the original BDD100K dataset, while YOLOP is expected to exhibit higher driving performance in rainy conditions when trained on our dataset with the RainSD module applied. Also, we discuss the association between the cloudiness and rainfall intensity on rainy road scenes and found a potential factor which can lead to a further development of rainy image synthesis. As future work, we plan to investigate on the aforementioned fascinating factor that will facilitate the rainy image generation, and expand the evaluation of our dataset to other multi-task networks for autonomus driving.

\section*{Acknowledgment}
This work was supported by the Korea Agency for Infrastructure Technology Advancement (KAIA) grant funded by the Ministry of Land, Infrastructure and Transport (21AMDP-C162419-01).

\bibliographystyle{elsarticle-num-names}
\bibliography{ref}

\begin{thebibliography}{55}
\expandafter\ifx\csname natexlab\endcsname\relax\def\natexlab#1{#1}\fi
\providecommand{\url}[1]{\texttt{#1}}
\providecommand{\href}[2]{#2}
\providecommand{\path}[1]{#1}
\providecommand{\DOIprefix}{doi:}
\providecommand{\ArXivprefix}{arXiv:}
\providecommand{\URLprefix}{URL: }
\providecommand{\Pubmedprefix}{pmid:}
\providecommand{\doi}[1]{\href{http://dx.doi.org/#1}{\path{#1}}}
\providecommand{\Pubmed}[1]{\href{pmid:#1}{\path{#1}}}
\providecommand{\bibinfo}[2]{#2}
\ifx\xfnm\relax \def\xfnm[#1]{\unskip,\space#1}\fi
%Type = Inproceedings
\bibitem[{Hauer et~al.(2019)Hauer, Schmidt, Holzm{\"u}ller, and Pretschner}]{1.hauer2019did}
\bibinfo{author}{F.~Hauer}, \bibinfo{author}{T.~Schmidt}, \bibinfo{author}{B.~Holzm{\"u}ller}, \bibinfo{author}{A.~Pretschner},
\newblock \bibinfo{title}{Did we test all scenarios for automated and autonomous driving systems?},
\newblock in: \bibinfo{booktitle}{2019 IEEE Intelligent Transportation Systems Conference (ITSC)}, \bibinfo{organization}{IEEE}, \bibinfo{year}{2019}, pp. \bibinfo{pages}{2950--2955}.
%Type = Inproceedings
\bibitem[{Ionita(2017)}]{2.ionita2017autonomous}
\bibinfo{author}{S.~Ionita},
\newblock \bibinfo{title}{Autonomous vehicles: from paradigms to technology},
\newblock in: \bibinfo{booktitle}{IOP Conference Series: Materials Science and Engineering}, volume \bibinfo{volume}{252}, \bibinfo{organization}{IOP Publishing}, \bibinfo{year}{2017}, p. \bibinfo{pages}{012098}.
%Type = Article
\bibitem[{Gkartzonikas and Gkritza(2019)}]{3.gkartzonikas2019have}
\bibinfo{author}{C.~Gkartzonikas}, \bibinfo{author}{K.~Gkritza},
\newblock \bibinfo{title}{What have we learned? a review of stated preference and choice studies on autonomous vehicles},
\newblock \bibinfo{journal}{Transportation Research Part C: Emerging Technologies} \bibinfo{volume}{98} (\bibinfo{year}{2019}) \bibinfo{pages}{323--337}.
%Type = Article
\bibitem[{Yeong et~al.(2021)Yeong, Velasco-Hernandez, Barry, and Walsh}]{4.yeong2021sensor}
\bibinfo{author}{D.~J. Yeong}, \bibinfo{author}{G.~Velasco-Hernandez}, \bibinfo{author}{J.~Barry}, \bibinfo{author}{J.~Walsh},
\newblock \bibinfo{title}{Sensor and sensor fusion technology in autonomous vehicles: A review},
\newblock \bibinfo{journal}{Sensors} \bibinfo{volume}{21} (\bibinfo{year}{2021}) \bibinfo{pages}{2140}.
%Type = Article
\bibitem[{Jing et~al.(2020)Jing, Xu, Chen, Shi, and Zhan}]{5.jing2020determinants}
\bibinfo{author}{P.~Jing}, \bibinfo{author}{G.~Xu}, \bibinfo{author}{Y.~Chen}, \bibinfo{author}{Y.~Shi}, \bibinfo{author}{F.~Zhan},
\newblock \bibinfo{title}{The determinants behind the acceptance of autonomous vehicles: A systematic review},
\newblock \bibinfo{journal}{Sustainability} \bibinfo{volume}{12} (\bibinfo{year}{2020}) \bibinfo{pages}{1719}.
%Type = Article
\bibitem[{Othman(2021)}]{6.othman2021public}
\bibinfo{author}{K.~Othman},
\newblock \bibinfo{title}{Public acceptance and perception of autonomous vehicles: a comprehensive review},
\newblock \bibinfo{journal}{AI and Ethics} \bibinfo{volume}{1} (\bibinfo{year}{2021}) \bibinfo{pages}{355--387}.
%Type = Article
\bibitem[{Cui et~al.(2019)Cui, Liew, Sabaliauskaite, and Zhou}]{7.cui2019review}
\bibinfo{author}{J.~Cui}, \bibinfo{author}{L.~S. Liew}, \bibinfo{author}{G.~Sabaliauskaite}, \bibinfo{author}{F.~Zhou},
\newblock \bibinfo{title}{A review on safety failures, security attacks, and available countermeasures for autonomous vehicles},
\newblock \bibinfo{journal}{Ad Hoc Networks} \bibinfo{volume}{90} (\bibinfo{year}{2019}) \bibinfo{pages}{101823}.
%Type = Article
\bibitem[{Hamzeh et~al.(2020)Hamzeh, El-Shair, and Rawashdeh}]{8.hamzeh2020effect}
\bibinfo{author}{Y.~Hamzeh}, \bibinfo{author}{Z.~El-Shair}, \bibinfo{author}{S.~A. Rawashdeh},
\newblock \bibinfo{title}{Effect of adherent rain on vision-based object detection algorithms},
\newblock \bibinfo{journal}{SAE International Journal of Advances and Current Practices in Mobility} \bibinfo{volume}{2} (\bibinfo{year}{2020}) \bibinfo{pages}{3051--3059}.
%Type = Article
\bibitem[{Hnewa and Radha(2020)}]{9.hnewa2020object}
\bibinfo{author}{M.~Hnewa}, \bibinfo{author}{H.~Radha},
\newblock \bibinfo{title}{Object detection under rainy conditions for autonomous vehicles: A review of state-of-the-art and emerging techniques},
\newblock \bibinfo{journal}{IEEE Signal Processing Magazine} \bibinfo{volume}{38} (\bibinfo{year}{2020}) \bibinfo{pages}{53--67}.
%Type = Article
\bibitem[{Goberville et~al.(2020)Goberville, El-Yabroudi, Omwanas, Rojas, Meyer, Asher, and Abdel-Qader}]{10.goberville2020analysis}
\bibinfo{author}{N.~Goberville}, \bibinfo{author}{M.~El-Yabroudi}, \bibinfo{author}{M.~Omwanas}, \bibinfo{author}{J.~Rojas}, \bibinfo{author}{R.~Meyer}, \bibinfo{author}{Z.~Asher}, \bibinfo{author}{I.~Abdel-Qader},
\newblock \bibinfo{title}{Analysis of lidar and camera data in real-world weather conditions for autonomous vehicle operations},
\newblock \bibinfo{journal}{SAE International Journal of Advances and Current Practices in Mobility} \bibinfo{volume}{2} (\bibinfo{year}{2020}) \bibinfo{pages}{2428--2434}.
%Type = Article
\bibitem[{Yoneda et~al.(2019)Yoneda, Suganuma, Yanase, and Aldibaja}]{11.yoneda2019automated}
\bibinfo{author}{K.~Yoneda}, \bibinfo{author}{N.~Suganuma}, \bibinfo{author}{R.~Yanase}, \bibinfo{author}{M.~Aldibaja},
\newblock \bibinfo{title}{Automated driving recognition technologies for adverse weather conditions},
\newblock \bibinfo{journal}{IATSS research} \bibinfo{volume}{43} (\bibinfo{year}{2019}) \bibinfo{pages}{253--262}.
%Type = Article
\bibitem[{Zio(2018)}]{30.zio2018future}
\bibinfo{author}{E.~Zio},
\newblock \bibinfo{title}{The future of risk assessment},
\newblock \bibinfo{journal}{Reliability Engineering \& System Safety} \bibinfo{volume}{177} (\bibinfo{year}{2018}) \bibinfo{pages}{176--190}.
%Type = Article
\bibitem[{Tang et~al.(2020)Tang, Shi, He, Sadek, and Qiao}]{12.tang2020performance}
\bibinfo{author}{L.~Tang}, \bibinfo{author}{Y.~Shi}, \bibinfo{author}{Q.~He}, \bibinfo{author}{A.~W. Sadek}, \bibinfo{author}{C.~Qiao},
\newblock \bibinfo{title}{Performance test of autonomous vehicle lidar sensors under different weather conditions},
\newblock \bibinfo{journal}{Transportation research record} \bibinfo{volume}{2674} (\bibinfo{year}{2020}) \bibinfo{pages}{319--329}.
%Type = Article
\bibitem[{Espineira et~al.(2021)Espineira, Robinson, Groenewald, Chan, and Donzella}]{13.espineira2021realistic}
\bibinfo{author}{J.~P. Espineira}, \bibinfo{author}{J.~Robinson}, \bibinfo{author}{J.~Groenewald}, \bibinfo{author}{P.~H. Chan}, \bibinfo{author}{V.~Donzella},
\newblock \bibinfo{title}{Realistic lidar with noise model for real-time testing of automated vehicles in a virtual environment},
\newblock \bibinfo{journal}{IEEE Sensors Journal} \bibinfo{volume}{21} (\bibinfo{year}{2021}) \bibinfo{pages}{9919--9926}.
%Type = Inproceedings
\bibitem[{Hoffman et~al.(2018)Hoffman, Tzeng, Park, Zhu, Isola, Saenko, Efros, and Darrell}]{31.hoffman2018cycada}
\bibinfo{author}{J.~Hoffman}, \bibinfo{author}{E.~Tzeng}, \bibinfo{author}{T.~Park}, \bibinfo{author}{J.-Y. Zhu}, \bibinfo{author}{P.~Isola}, \bibinfo{author}{K.~Saenko}, \bibinfo{author}{A.~Efros}, \bibinfo{author}{T.~Darrell},
\newblock \bibinfo{title}{Cycada: Cycle-consistent adversarial domain adaptation},
\newblock in: \bibinfo{booktitle}{International conference on machine learning}, \bibinfo{organization}{Pmlr}, \bibinfo{year}{2018}, pp. \bibinfo{pages}{1989--1998}.
%Type = Article
\bibitem[{Liu et~al.(2017)Liu, Breuel, and Kautz}]{32.liu2017unsupervised}
\bibinfo{author}{M.-Y. Liu}, \bibinfo{author}{T.~Breuel}, \bibinfo{author}{J.~Kautz},
\newblock \bibinfo{title}{Unsupervised image-to-image translation networks},
\newblock \bibinfo{journal}{Advances in neural information processing systems} \bibinfo{volume}{30} (\bibinfo{year}{2017}).
%Type = Inproceedings
\bibitem[{Zheng et~al.(2020)Zheng, Wu, Han, and Shi}]{33.zheng2020forkgan}
\bibinfo{author}{Z.~Zheng}, \bibinfo{author}{Y.~Wu}, \bibinfo{author}{X.~Han}, \bibinfo{author}{J.~Shi},
\newblock \bibinfo{title}{Forkgan: Seeing into the rainy night},
\newblock in: \bibinfo{booktitle}{Computer Vision--ECCV 2020: 16th European Conference, Glasgow, UK, August 23--28, 2020, Proceedings, Part III 16}, \bibinfo{organization}{Springer}, \bibinfo{year}{2020}, pp. \bibinfo{pages}{155--170}.
%Type = Article
\bibitem[{Zhu et~al.(2017)Zhu, Zhang, Pathak, Darrell, Efros, Wang, and Shechtman}]{34.zhu2017toward}
\bibinfo{author}{J.-Y. Zhu}, \bibinfo{author}{R.~Zhang}, \bibinfo{author}{D.~Pathak}, \bibinfo{author}{T.~Darrell}, \bibinfo{author}{A.~A. Efros}, \bibinfo{author}{O.~Wang}, \bibinfo{author}{E.~Shechtman},
\newblock \bibinfo{title}{Toward multimodal image-to-image translation},
\newblock \bibinfo{journal}{Advances in neural information processing systems} \bibinfo{volume}{30} (\bibinfo{year}{2017}).
%Type = Inproceedings
\bibitem[{Xia et~al.(2023)Xia, Monica, Chao, Hariharan, Weinberger, and Campbell}]{35.xia2023image}
\bibinfo{author}{Y.~Xia}, \bibinfo{author}{J.~Monica}, \bibinfo{author}{W.-L. Chao}, \bibinfo{author}{B.~Hariharan}, \bibinfo{author}{K.~Q. Weinberger}, \bibinfo{author}{M.~Campbell},
\newblock \bibinfo{title}{Image-to-image translation for autonomous driving from coarsely-aligned image pairs},
\newblock in: \bibinfo{booktitle}{2023 IEEE International Conference on Robotics and Automation (ICRA)}, \bibinfo{organization}{IEEE}, \bibinfo{year}{2023}, pp. \bibinfo{pages}{7756--7762}.
%Type = Article
\bibitem[{Schutera et~al.(2020)Schutera, Hussein, Abhau, Mikut, and Reischl}]{36.schutera2020night}
\bibinfo{author}{M.~Schutera}, \bibinfo{author}{M.~Hussein}, \bibinfo{author}{J.~Abhau}, \bibinfo{author}{R.~Mikut}, \bibinfo{author}{M.~Reischl},
\newblock \bibinfo{title}{Night-to-day: Online image-to-image translation for object detection within autonomous driving by night},
\newblock \bibinfo{journal}{IEEE Transactions on Intelligent Vehicles} \bibinfo{volume}{6} (\bibinfo{year}{2020}) \bibinfo{pages}{480--489}.
%Type = Inproceedings
\bibitem[{Arruda et~al.(2019)Arruda, Paixao, Berriel, De~Souza, Badue, Sebe, and Oliveira-Santos}]{37.arruda2019cross}
\bibinfo{author}{V.~F. Arruda}, \bibinfo{author}{T.~M. Paixao}, \bibinfo{author}{R.~F. Berriel}, \bibinfo{author}{A.~F. De~Souza}, \bibinfo{author}{C.~Badue}, \bibinfo{author}{N.~Sebe}, \bibinfo{author}{T.~Oliveira-Santos},
\newblock \bibinfo{title}{Cross-domain car detection using unsupervised image-to-image translation: From day to night},
\newblock in: \bibinfo{booktitle}{2019 International Joint Conference on Neural Networks (IJCNN)}, \bibinfo{organization}{IEEE}, \bibinfo{year}{2019}, pp. \bibinfo{pages}{1--8}.
%Type = Article
\bibitem[{Lin et~al.(2020)Lin, Huang, Wu, and Lai}]{38.lin2020gan}
\bibinfo{author}{C.-T. Lin}, \bibinfo{author}{S.-W. Huang}, \bibinfo{author}{Y.-Y. Wu}, \bibinfo{author}{S.-H. Lai},
\newblock \bibinfo{title}{Gan-based day-to-night image style transfer for nighttime vehicle detection},
\newblock \bibinfo{journal}{IEEE Transactions on Intelligent Transportation Systems} \bibinfo{volume}{22} (\bibinfo{year}{2020}) \bibinfo{pages}{951--963}.
%Type = Inproceedings
\bibitem[{Jin et~al.(2021)Jin, Fatemi, Lira, Yu, Leng, Ma, Mahdavi-Amiri, and Zhang}]{39.jin2021raidar}
\bibinfo{author}{J.~Jin}, \bibinfo{author}{A.~Fatemi}, \bibinfo{author}{W.~M.~P. Lira}, \bibinfo{author}{F.~Yu}, \bibinfo{author}{B.~Leng}, \bibinfo{author}{R.~Ma}, \bibinfo{author}{A.~Mahdavi-Amiri}, \bibinfo{author}{H.~Zhang},
\newblock \bibinfo{title}{Raidar: A rich annotated image dataset of rainy street scenes},
\newblock in: \bibinfo{booktitle}{Proceedings of the IEEE/CVF International Conference on Computer Vision}, \bibinfo{year}{2021}, pp. \bibinfo{pages}{2951--2961}.
%Type = Inproceedings
\bibitem[{Jeong et~al.(2021)Jeong, Kim, Lee, and Sohn}]{40.jeong2021memory}
\bibinfo{author}{S.~Jeong}, \bibinfo{author}{Y.~Kim}, \bibinfo{author}{E.~Lee}, \bibinfo{author}{K.~Sohn},
\newblock \bibinfo{title}{Memory-guided unsupervised image-to-image translation},
\newblock in: \bibinfo{booktitle}{Proceedings of the IEEE/CVF conference on computer vision and pattern recognition}, \bibinfo{year}{2021}, pp. \bibinfo{pages}{6558--6567}.
%Type = Inproceedings
\bibitem[{Yu et~al.(2020)Yu, Chen, Wang, Xian, Chen, Liu, Madhavan, and Darrell}]{41.yu2020bdd100k}
\bibinfo{author}{F.~Yu}, \bibinfo{author}{H.~Chen}, \bibinfo{author}{X.~Wang}, \bibinfo{author}{W.~Xian}, \bibinfo{author}{Y.~Chen}, \bibinfo{author}{F.~Liu}, \bibinfo{author}{V.~Madhavan}, \bibinfo{author}{T.~Darrell},
\newblock \bibinfo{title}{Bdd100k: A diverse driving dataset for heterogeneous multitask learning},
\newblock in: \bibinfo{booktitle}{Proceedings of the IEEE/CVF conference on computer vision and pattern recognition}, \bibinfo{year}{2020}, pp. \bibinfo{pages}{2636--2645}.
%Type = Inproceedings
\bibitem[{Jiang et~al.(2020)Jiang, Zhang, Huang, Liu, Shi, and Loy}]{42.jiang2020tsit}
\bibinfo{author}{L.~Jiang}, \bibinfo{author}{C.~Zhang}, \bibinfo{author}{M.~Huang}, \bibinfo{author}{C.~Liu}, \bibinfo{author}{J.~Shi}, \bibinfo{author}{C.~C. Loy},
\newblock \bibinfo{title}{Tsit: A simple and versatile framework for image-to-image translation},
\newblock in: \bibinfo{booktitle}{Computer Vision--ECCV 2020: 16th European Conference, Glasgow, UK, August 23--28, 2020, Proceedings, Part III 16}, \bibinfo{organization}{Springer}, \bibinfo{year}{2020}, pp. \bibinfo{pages}{206--222}.
%Type = Inproceedings
\bibitem[{Cordts et~al.(2016)Cordts, Omran, Ramos, Rehfeld, Enzweiler, Benenson, Franke, Roth, and Schiele}]{43.cordts2016cityscapes}
\bibinfo{author}{M.~Cordts}, \bibinfo{author}{M.~Omran}, \bibinfo{author}{S.~Ramos}, \bibinfo{author}{T.~Rehfeld}, \bibinfo{author}{M.~Enzweiler}, \bibinfo{author}{R.~Benenson}, \bibinfo{author}{U.~Franke}, \bibinfo{author}{S.~Roth}, \bibinfo{author}{B.~Schiele},
\newblock \bibinfo{title}{The cityscapes dataset for semantic urban scene understanding},
\newblock in: \bibinfo{booktitle}{Proceedings of the IEEE conference on computer vision and pattern recognition}, \bibinfo{year}{2016}, pp. \bibinfo{pages}{3213--3223}.
%Type = Article
\bibitem[{Geiger et~al.(2013)Geiger, Lenz, Stiller, and Urtasun}]{44.geiger2013vision}
\bibinfo{author}{A.~Geiger}, \bibinfo{author}{P.~Lenz}, \bibinfo{author}{C.~Stiller}, \bibinfo{author}{R.~Urtasun},
\newblock \bibinfo{title}{Vision meets robotics: The kitti dataset},
\newblock \bibinfo{journal}{The International Journal of Robotics Research} \bibinfo{volume}{32} (\bibinfo{year}{2013}) \bibinfo{pages}{1231--1237}.
%Type = Inproceedings
\bibitem[{Caesar et~al.(2020)Caesar, Bankiti, Lang, Vora, Liong, Xu, Krishnan, Pan, Baldan, and Beijbom}]{45.caesar2020nuscenes}
\bibinfo{author}{H.~Caesar}, \bibinfo{author}{V.~Bankiti}, \bibinfo{author}{A.~H. Lang}, \bibinfo{author}{S.~Vora}, \bibinfo{author}{V.~E. Liong}, \bibinfo{author}{Q.~Xu}, \bibinfo{author}{A.~Krishnan}, \bibinfo{author}{Y.~Pan}, \bibinfo{author}{G.~Baldan}, \bibinfo{author}{O.~Beijbom},
\newblock \bibinfo{title}{nuscenes: A multimodal dataset for autonomous driving},
\newblock in: \bibinfo{booktitle}{Proceedings of the IEEE/CVF conference on computer vision and pattern recognition}, \bibinfo{year}{2020}, pp. \bibinfo{pages}{11621--11631}.
%Type = Inproceedings
\bibitem[{Ros et~al.(2016)Ros, Sellart, Materzynska, Vazquez, and Lopez}]{14.ros2016synthia}
\bibinfo{author}{G.~Ros}, \bibinfo{author}{L.~Sellart}, \bibinfo{author}{J.~Materzynska}, \bibinfo{author}{D.~Vazquez}, \bibinfo{author}{A.~M. Lopez},
\newblock \bibinfo{title}{The synthia dataset: A large collection of synthetic images for semantic segmentation of urban scenes},
\newblock in: \bibinfo{booktitle}{Proceedings of the IEEE conference on computer vision and pattern recognition}, \bibinfo{year}{2016}, pp. \bibinfo{pages}{3234--3243}.
%Type = Article
\bibitem[{Wrenninge and Unger(2018)}]{15.wrenninge2018synscapes}
\bibinfo{author}{M.~Wrenninge}, \bibinfo{author}{J.~Unger},
\newblock \bibinfo{title}{Synscapes: A photorealistic synthetic dataset for street scene parsing},
\newblock \bibinfo{journal}{arXiv preprint arXiv:1810.08705}  (\bibinfo{year}{2018}).
%Type = Inproceedings
\bibitem[{Dosovitskiy et~al.(2017)Dosovitskiy, Ros, Codevilla, Lopez, and Koltun}]{16.dosovitskiy2017carla}
\bibinfo{author}{A.~Dosovitskiy}, \bibinfo{author}{G.~Ros}, \bibinfo{author}{F.~Codevilla}, \bibinfo{author}{A.~Lopez}, \bibinfo{author}{V.~Koltun},
\newblock \bibinfo{title}{Carla: An open urban driving simulator},
\newblock in: \bibinfo{booktitle}{Conference on robot learning}, \bibinfo{organization}{PMLR}, \bibinfo{year}{2017}, pp. \bibinfo{pages}{1--16}.
%Type = Inproceedings
\bibitem[{Rong et~al.(2020)Rong, Shin, Tabatabaee, Lu, Lemke, Mo{\v{z}}eiko, Boise, Uhm, Gerow, Mehta et~al.}]{17.rong2020lgsvl}
\bibinfo{author}{G.~Rong}, \bibinfo{author}{B.~H. Shin}, \bibinfo{author}{H.~Tabatabaee}, \bibinfo{author}{Q.~Lu}, \bibinfo{author}{S.~Lemke}, \bibinfo{author}{M.~Mo{\v{z}}eiko}, \bibinfo{author}{E.~Boise}, \bibinfo{author}{G.~Uhm}, \bibinfo{author}{M.~Gerow}, \bibinfo{author}{S.~Mehta}, et~al.,
\newblock \bibinfo{title}{Lgsvl simulator: A high fidelity simulator for autonomous driving},
\newblock in: \bibinfo{booktitle}{2020 IEEE 23rd International conference on intelligent transportation systems (ITSC)}, \bibinfo{organization}{IEEE}, \bibinfo{year}{2020}, pp. \bibinfo{pages}{1--6}.
%Type = Inproceedings
\bibitem[{Shah et~al.(2018)Shah, Dey, Lovett, and Kapoor}]{18.shah2018airsim}
\bibinfo{author}{S.~Shah}, \bibinfo{author}{D.~Dey}, \bibinfo{author}{C.~Lovett}, \bibinfo{author}{A.~Kapoor},
\newblock \bibinfo{title}{Airsim: High-fidelity visual and physical simulation for autonomous vehicles},
\newblock in: \bibinfo{booktitle}{Field and Service Robotics: Results of the 11th International Conference}, \bibinfo{organization}{Springer}, \bibinfo{year}{2018}, pp. \bibinfo{pages}{621--635}.
%Type = Article
\bibitem[{Jeon et~al.(2022)Jeon, Kim, Choi, Park, Son, Lee, Choi, and Lim}]{19.jeon2022carla}
\bibinfo{author}{H.~Jeon}, \bibinfo{author}{Y.~Kim}, \bibinfo{author}{M.~Choi}, \bibinfo{author}{D.~Park}, \bibinfo{author}{S.~Son}, \bibinfo{author}{J.~Lee}, \bibinfo{author}{G.~Choi}, \bibinfo{author}{Y.~Lim},
\newblock \bibinfo{title}{Carla simulator-based evaluation framework development of lane detection accuracy performance under sensor blockage caused by heavy rain for autonomous vehicle},
\newblock \bibinfo{journal}{IEEE Robotics and Automation Letters} \bibinfo{volume}{7} (\bibinfo{year}{2022}) \bibinfo{pages}{9977--9984}.
%Type = Inproceedings
\bibitem[{Zhu et~al.(2019)Zhu, Sapra, Reda, Shih, Newsam, Tao, and Catanzaro}]{48.zhu2019improving}
\bibinfo{author}{Y.~Zhu}, \bibinfo{author}{K.~Sapra}, \bibinfo{author}{F.~A. Reda}, \bibinfo{author}{K.~J. Shih}, \bibinfo{author}{S.~Newsam}, \bibinfo{author}{A.~Tao}, \bibinfo{author}{B.~Catanzaro},
\newblock \bibinfo{title}{Improving semantic segmentation via video propagation and label relaxation},
\newblock in: \bibinfo{booktitle}{Proceedings of the IEEE/CVF conference on computer vision and pattern recognition}, \bibinfo{year}{2019}, pp. \bibinfo{pages}{8856--8865}.
%Type = Article
\bibitem[{Tao et~al.(2020)Tao, Sapra, and Catanzaro}]{49.tao2020hierarchical}
\bibinfo{author}{A.~Tao}, \bibinfo{author}{K.~Sapra}, \bibinfo{author}{B.~Catanzaro},
\newblock \bibinfo{title}{Hierarchical multi-scale attention for semantic segmentation},
\newblock \bibinfo{journal}{arXiv preprint arXiv:2005.10821}  (\bibinfo{year}{2020}).
%Type = Inproceedings
\bibitem[{Kirillov et~al.(2020)Kirillov, Wu, He, and Girshick}]{50.kirillov2020pointrend}
\bibinfo{author}{A.~Kirillov}, \bibinfo{author}{Y.~Wu}, \bibinfo{author}{K.~He}, \bibinfo{author}{R.~Girshick},
\newblock \bibinfo{title}{Pointrend: Image segmentation as rendering},
\newblock in: \bibinfo{booktitle}{Proceedings of the IEEE/CVF conference on computer vision and pattern recognition}, \bibinfo{year}{2020}, pp. \bibinfo{pages}{9799--9808}.
%Type = Inproceedings
\bibitem[{Chen et~al.(2019)Chen, Girshick, He, and Doll{\'a}r}]{51.chen2019tensormask}
\bibinfo{author}{X.~Chen}, \bibinfo{author}{R.~Girshick}, \bibinfo{author}{K.~He}, \bibinfo{author}{P.~Doll{\'a}r},
\newblock \bibinfo{title}{Tensormask: A foundation for dense object segmentation},
\newblock in: \bibinfo{booktitle}{Proceedings of the IEEE/CVF international conference on computer vision}, \bibinfo{year}{2019}, pp. \bibinfo{pages}{2061--2069}.
%Type = Article
\bibitem[{Pang et~al.(2021)Pang, Lin, Qin, and Chen}]{47.pang2021image}
\bibinfo{author}{Y.~Pang}, \bibinfo{author}{J.~Lin}, \bibinfo{author}{T.~Qin}, \bibinfo{author}{Z.~Chen},
\newblock \bibinfo{title}{Image-to-image translation: Methods and applications},
\newblock \bibinfo{journal}{IEEE Transactions on Multimedia} \bibinfo{volume}{24} (\bibinfo{year}{2021}) \bibinfo{pages}{3859--3881}.
%Type = Inproceedings
\bibitem[{Lee et~al.(2018)Lee, Tseng, Huang, Singh, and Yang}]{53.lee2018diverse}
\bibinfo{author}{H.-Y. Lee}, \bibinfo{author}{H.-Y. Tseng}, \bibinfo{author}{J.-B. Huang}, \bibinfo{author}{M.~Singh}, \bibinfo{author}{M.-H. Yang},
\newblock \bibinfo{title}{Diverse image-to-image translation via disentangled representations},
\newblock in: \bibinfo{booktitle}{Proceedings of the European conference on computer vision (ECCV)}, \bibinfo{year}{2018}, pp. \bibinfo{pages}{35--51}.
%Type = Article
\bibitem[{Pahk et~al.(2023)Pahk, Shim, Baek, Lim, and Choi}]{54.pahk2023effects}
\bibinfo{author}{J.~Pahk}, \bibinfo{author}{J.~Shim}, \bibinfo{author}{M.~Baek}, \bibinfo{author}{Y.~Lim}, \bibinfo{author}{G.~Choi},
\newblock \bibinfo{title}{Effects of sim2real image translation via dclgan on lane keeping assist system in carla simulator},
\newblock \bibinfo{journal}{IEEE Access} \bibinfo{volume}{11} (\bibinfo{year}{2023}) \bibinfo{pages}{33915--33927}.
%Type = Inproceedings
\bibitem[{Pizzati et~al.(2020)Pizzati, Charette, Zaccaria, and Cerri}]{52.pizzati2020domain}
\bibinfo{author}{F.~Pizzati}, \bibinfo{author}{R.~d. Charette}, \bibinfo{author}{M.~Zaccaria}, \bibinfo{author}{P.~Cerri},
\newblock \bibinfo{title}{Domain bridge for unpaired image-to-image translation and unsupervised domain adaptation},
\newblock in: \bibinfo{booktitle}{Proceedings of the IEEE/CVF winter conference on applications of computer vision}, \bibinfo{year}{2020}, pp. \bibinfo{pages}{2990--2998}.
%Type = Inproceedings
\bibitem[{Isola et~al.(2017)Isola, Zhu, Zhou, and Efros}]{55.isola2017image}
\bibinfo{author}{P.~Isola}, \bibinfo{author}{J.-Y. Zhu}, \bibinfo{author}{T.~Zhou}, \bibinfo{author}{A.~A. Efros},
\newblock \bibinfo{title}{Image-to-image translation with conditional adversarial networks},
\newblock in: \bibinfo{booktitle}{Proceedings of the IEEE conference on computer vision and pattern recognition}, \bibinfo{year}{2017}, pp. \bibinfo{pages}{1125--1134}.
%Type = Inproceedings
\bibitem[{Zhu et~al.(2017)Zhu, Park, Isola, and Efros}]{56.zhu2017unpaired}
\bibinfo{author}{J.-Y. Zhu}, \bibinfo{author}{T.~Park}, \bibinfo{author}{P.~Isola}, \bibinfo{author}{A.~A. Efros},
\newblock \bibinfo{title}{Unpaired image-to-image translation using cycle-consistent adversarial networks},
\newblock in: \bibinfo{booktitle}{Proceedings of the IEEE international conference on computer vision}, \bibinfo{year}{2017}, pp. \bibinfo{pages}{2223--2232}.
%Type = Inproceedings
\bibitem[{Huang et~al.(2018)Huang, Liu, Belongie, and Kautz}]{57.huang2018multimodal}
\bibinfo{author}{X.~Huang}, \bibinfo{author}{M.-Y. Liu}, \bibinfo{author}{S.~Belongie}, \bibinfo{author}{J.~Kautz},
\newblock \bibinfo{title}{Multimodal unsupervised image-to-image translation},
\newblock in: \bibinfo{booktitle}{Proceedings of the European conference on computer vision (ECCV)}, \bibinfo{year}{2018}, pp. \bibinfo{pages}{172--189}.
%Type = Article
\bibitem[{Lee et~al.(2020)Lee, Tseng, Mao, Huang, Lu, Singh, and Yang}]{58.lee2020drit++}
\bibinfo{author}{H.-Y. Lee}, \bibinfo{author}{H.-Y. Tseng}, \bibinfo{author}{Q.~Mao}, \bibinfo{author}{J.-B. Huang}, \bibinfo{author}{Y.-D. Lu}, \bibinfo{author}{M.~Singh}, \bibinfo{author}{M.-H. Yang},
\newblock \bibinfo{title}{Drit++: Diverse image-to-image translation via disentangled representations},
\newblock \bibinfo{journal}{International Journal of Computer Vision} \bibinfo{volume}{128} (\bibinfo{year}{2020}) \bibinfo{pages}{2402--2417}.
%Type = Article
\bibitem[{Chen et~al.(2018)Chen, Zhao, Lv, and Zhang}]{60.chen2018multi}
\bibinfo{author}{Y.~Chen}, \bibinfo{author}{D.~Zhao}, \bibinfo{author}{L.~Lv}, \bibinfo{author}{Q.~Zhang},
\newblock \bibinfo{title}{Multi-task learning for dangerous object detection in autonomous driving},
\newblock \bibinfo{journal}{Information Sciences} \bibinfo{volume}{432} (\bibinfo{year}{2018}) \bibinfo{pages}{559--571}.
%Type = Article
\bibitem[{Zhang and Yang(2021)}]{61.zhang2021survey}
\bibinfo{author}{Y.~Zhang}, \bibinfo{author}{Q.~Yang},
\newblock \bibinfo{title}{A survey on multi-task learning},
\newblock \bibinfo{journal}{IEEE Transactions on Knowledge and Data Engineering} \bibinfo{volume}{34} (\bibinfo{year}{2021}) \bibinfo{pages}{5586--5609}.
%Type = Inproceedings
\bibitem[{Liu et~al.(2019)Liu, Johns, and Davison}]{62.liu2019end}
\bibinfo{author}{S.~Liu}, \bibinfo{author}{E.~Johns}, \bibinfo{author}{A.~J. Davison},
\newblock \bibinfo{title}{End-to-end multi-task learning with attention},
\newblock in: \bibinfo{booktitle}{Proceedings of the IEEE/CVF conference on computer vision and pattern recognition}, \bibinfo{year}{2019}, pp. \bibinfo{pages}{1871--1880}.
%Type = Article
\bibitem[{Wu et~al.(2022)Wu, Liao, Zhang, Wang, Bai, Cheng, and Liu}]{63.wu2022yolop}
\bibinfo{author}{D.~Wu}, \bibinfo{author}{M.-W. Liao}, \bibinfo{author}{W.-T. Zhang}, \bibinfo{author}{X.-G. Wang}, \bibinfo{author}{X.~Bai}, \bibinfo{author}{W.-Q. Cheng}, \bibinfo{author}{W.-Y. Liu},
\newblock \bibinfo{title}{Yolop: You only look once for panoptic driving perception},
\newblock \bibinfo{journal}{Machine Intelligence Research} \bibinfo{volume}{19} (\bibinfo{year}{2022}) \bibinfo{pages}{550--562}.
%Type = Article
\bibitem[{Vu et~al.(2022)Vu, Ngo, and Phan}]{64.vu2022hybridnets}
\bibinfo{author}{D.~Vu}, \bibinfo{author}{B.~Ngo}, \bibinfo{author}{H.~Phan},
\newblock \bibinfo{title}{Hybridnets: End-to-end perception network},
\newblock \bibinfo{journal}{arXiv preprint arXiv:2203.09035}  (\bibinfo{year}{2022}).
%Type = Inproceedings
\bibitem[{Johnson et~al.(2016)Johnson, Alahi, and Fei-Fei}]{65.johnson2016perceptual}
\bibinfo{author}{J.~Johnson}, \bibinfo{author}{A.~Alahi}, \bibinfo{author}{L.~Fei-Fei},
\newblock \bibinfo{title}{Perceptual losses for real-time style transfer and super-resolution},
\newblock in: \bibinfo{booktitle}{Computer Vision--ECCV 2016: 14th European Conference, Amsterdam, The Netherlands, October 11-14, 2016, Proceedings, Part II 14}, \bibinfo{organization}{Springer}, \bibinfo{year}{2016}, pp. \bibinfo{pages}{694--711}.
%Type = Inproceedings
\bibitem[{Wang et~al.(2018)Wang, Liu, Zhu, Tao, Kautz, and Catanzaro}]{66.wang2018high}
\bibinfo{author}{T.-C. Wang}, \bibinfo{author}{M.-Y. Liu}, \bibinfo{author}{J.-Y. Zhu}, \bibinfo{author}{A.~Tao}, \bibinfo{author}{J.~Kautz}, \bibinfo{author}{B.~Catanzaro},
\newblock \bibinfo{title}{High-resolution image synthesis and semantic manipulation with conditional gans},
\newblock in: \bibinfo{booktitle}{Proceedings of the IEEE conference on computer vision and pattern recognition}, \bibinfo{year}{2018}, pp. \bibinfo{pages}{8798--8807}.
%Type = Article
\bibitem[{Paszke et~al.(2019)Paszke, Gross, Massa, Lerer, Bradbury, Chanan, Killeen, Lin, Gimelshein, Antiga et~al.}]{67.paszke2019pytorch}
\bibinfo{author}{A.~Paszke}, \bibinfo{author}{S.~Gross}, \bibinfo{author}{F.~Massa}, \bibinfo{author}{A.~Lerer}, \bibinfo{author}{J.~Bradbury}, \bibinfo{author}{G.~Chanan}, \bibinfo{author}{T.~Killeen}, \bibinfo{author}{Z.~Lin}, \bibinfo{author}{N.~Gimelshein}, \bibinfo{author}{L.~Antiga}, et~al.,
\newblock \bibinfo{title}{Pytorch: An imperative style, high-performance deep learning library},
\newblock \bibinfo{journal}{Advances in neural information processing systems} \bibinfo{volume}{32} (\bibinfo{year}{2019}).

\end{thebibliography}
\clearpage
\setlength\intextsep{0pt}
\begin{wrapfigure}{l}{20mm} 
\includegraphics[width=1in,clip,keepaspectratio]{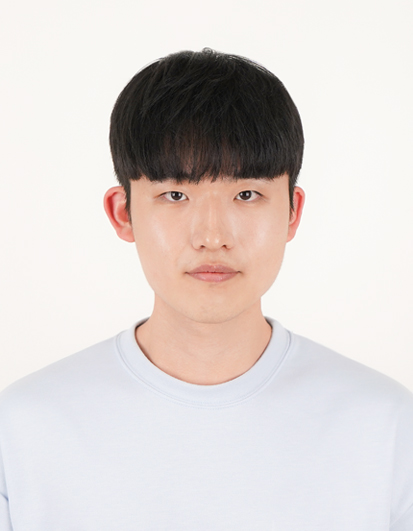}
\end{wrapfigure}\par
\noindent \textbf{Hyeonjae Jeon} is a graduate student in the Robotics and Mechatronics Engineering Department at Daegu Gyeongbuk Institute of Science and Technology (DGIST) in South Korea. He received his B.S from DGIST in 2021. His research interests include autonomous vehicles, computer vision, domain adaptation, feature disentanglement, and multi-task learning. 
\vspace{-0.7cm}
\subsection*{  }
\subsection*{  }

\setlength\intextsep{0pt}
\begin{wrapfigure}{l}{20mm} 
\includegraphics[width=1in,clip,keepaspectratio]{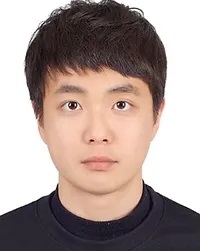}
\end{wrapfigure}\par
\noindent \textbf{Junghyun Seo} is a graduate student in the Robotics and Mechatronics Engineering Department at DGIST in South Korea. His research interests include autonomous vehicles, computer vision, multi-modal learning. 
\vspace{-0.5cm}
\subsection*{  }
\subsection*{  }

\setlength\intextsep{0pt}
\begin{wrapfigure}{l}{20mm} 
\includegraphics[width=1in,clip,keepaspectratio]{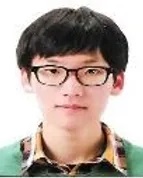}
\end{wrapfigure}\par
\noindent \textbf{Taesoo Kim} is a graduate student in the Robotics and Mechatronics Engineering Department at DGIST in South Korea. He received his B.S from DGIST in 2022. His research interests include autonomous vehicles, computer vision, and image-to-image translation. 
\vspace{-0.5cm}

\subsection*{  }

\setlength\intextsep{0pt}
\begin{wrapfigure}{l}{20mm} 
\includegraphics[width=1in,clip,keepaspectratio]{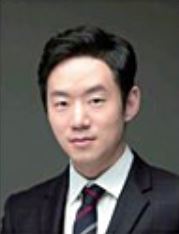}
\end{wrapfigure}\par
\noindent \textbf{Sungho Son} is a Principal Researcher at Korea Automotive Testing and Research Institute(KATRI). His research interests include autonomous vehicles, sensor blockage, and sensor cleaning. 

\subsection*{  }
\subsection*{  }
\subsection*{  }

\setlength\intextsep{0pt}
\begin{wrapfigure}{l}{20mm} 
\includegraphics[width=1in,clip,keepaspectratio]{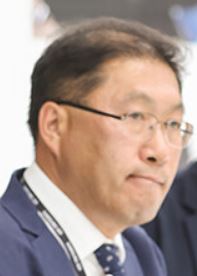}
\end{wrapfigure}\par
\noindent \textbf{Jungki Lee} Jungki Lee is the Head of the Autonomous Driving Division at Korea Automotive Testing and Research Institute(KATRI). His research interests include autonomous vehicles, sensor blockage, and sensor cleaning.
% \vspace{-0.5cm}
\subsection*{  }
\subsection*{  }
\setlength\intextsep{0pt}
\begin{wrapfigure}{l}{20mm} 
\includegraphics[width=1in,clip,keepaspectratio]{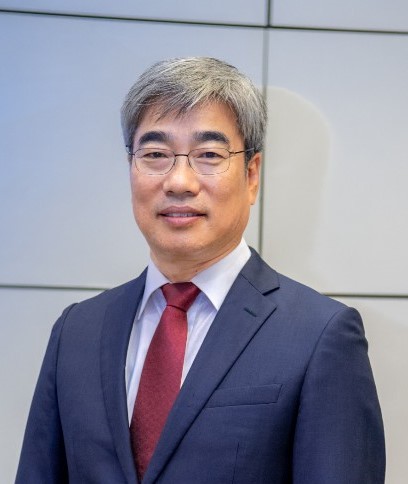}
\end{wrapfigure}\par
\noindent \textbf{Gyeungho Choi} received the Ph.D. degree in mechanical engineering from The University of Alabama, Tuscaloosa, AL, USA, in 1992, and the Ph.D. degree (Hons.) in automotive and energy engineering conferred by His Royal Highness Princess Maha Chakri Sirindhorn from the King Mongkut’s University of Technology North Bangkok, Thailand, in 2018. He has taught at Keimyung University and the Daegu Gyeongbuk Institute of Science and Technology (DGIST), as a Professor, for 30 years. His research interests include self-driving vehicles, ADAS, vehicle-in-the-loop systems, and alternative energy utilization. He served as the President for Korean Auto-Vehicle Safety Association and the Editor-in-Chief for Korean Society for Engineering Education.
\subsection*{  }

\setlength\intextsep{0pt}
\begin{wrapfigure}{l}{20mm} 
\includegraphics[width=1in,clip,keepaspectratio]{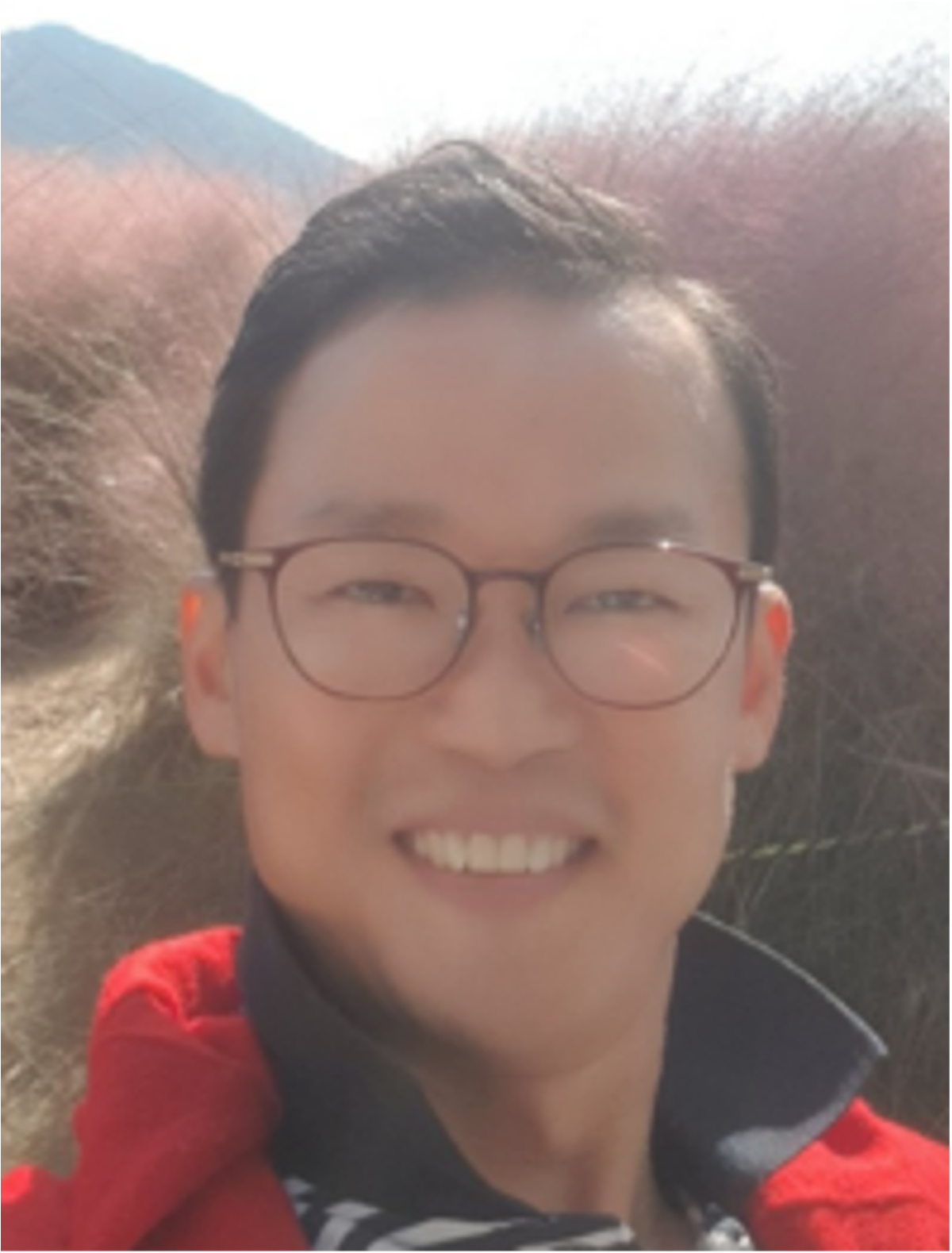}
\end{wrapfigure}\par
\noindent \textbf{Yongseob Lim} received the Ph.D. degree in mechanical engineering from the University of Michigan, Ann Arbor, MI, USA, in 2010. He was with the Hyundai Motor Company R\&D Advanced Chassis Platform Research Group, South Korea, as a Research Engineer. He was also with Samsung Techwin Company Mechatronics System Development Group, South Korea as a Principal Research Engineer. He is currently an Associate Professor in the Robotics Engineering Department, Daegu Gyeongbuk Institute of Science and Technology (DGIST), and the Co-Director of the Joint Research Laboratory of Autonomous Systems and Control (ASC) and Vehicle in the Loop Simulation (VILS) Laboratories. His research interests include the modeling, control, and design of mechatronics systems with particular interests in autonomous ground vehicles and flight robotics systems and control. \par

\end{document}